\def\forarxiv{1}
\def\foricml{0} 
\def\foreaamo{0} 

\if\forarxiv 1
\documentclass{article}
\everypar{\looseness=-1}
\usepackage{fullpage}
\fi 

\if\foreaamo 1
\documentclass{article}
\everypar{\looseness=-1}
\usepackage{fullpage}
\usepackage{PRIMEarxiv}
\fi

\usepackage{eec227c}
\usepackage{authblk}
\usepackage{microtype}
\usepackage{graphicx}
\usepackage{subfigure}
\usepackage{booktabs} 

\usepackage{hyperref}

\usepackage{algorithm}
\usepackage{algorithmic}

\usepackage{amsmath}
\usepackage{mathtools}
\usepackage{graphicx} 
\usepackage{shortcuts} 
\usepackage{latexsym}
\usepackage{amscd}
\usepackage{amsbsy}
\usepackage{amssymb}
\usepackage{float}
\usepackage{amsthm}
\usepackage{comment} 
\usepackage{cleveref}
\usepackage{nicefrac}

\newtheorem{theorem}{Theorem}
\newtheorem{lemma}{Lemma}
\newtheorem{corollary}[theorem]{Corollary}
\theoremstyle{definition}
\newtheorem{assumption}{Assumption}
\newtheorem{definition}{Definition}
\newtheorem{proposition}{Proposition}

\crefname{assumption}{assumption}{assumptions}

\usepackage{xcolor}
\usepackage{url}
\usepackage{natbib}
\if\foreaamo 1
\fi

\theoremstyle{plain}

\usepackage[textsize=tiny]{todonotes}

\begin{document}

\if\foricml 1
\twocolumn[
\icmltitle{Reduced-Rank Multi-objective Policy Learning and Optimization}

\begin{icmlauthorlist}
\icmlauthor{Ezinne Nwankwo}{eecs_berkeley}
\icmlauthor{Michael I. Jordan}{eecs_berkeley}
\icmlauthor{Angela Zhou}{ds_usc}
\end{icmlauthorlist}

\icmlaffiliation{eecs_berkeley}{Department of Electrical Engineering and Computer Sciences, University of California, Berkeley, USA}
\icmlaffiliation{ds_usc}{Department of Data Sciences and Operations Research, University of Southern California, USA}

\icmlcorrespondingauthor{Ezinne Nwankwo}{ezinne\_nwankwo@berkeley.edu}

\vskip 0.3in
]

\printAffiliationsAndNotice{} 
\fi 
\if \forarxiv 1
\title{Reduced-Rank Multi-objective Policy Learning and Optimization}
\author[1]{Ezinne Nwankwo}
\author[1,2]{Michael I. Jordan}
\author[3]{Angela Zhou}
\affil[1]{Department of Electrical Engineering and Computer Sciences, University of California, Berkeley}
\affil[2]{Department of Statistics, University of California, Berkeley}
\affil[3]{Department of Data Sciences and Operations, University of Southern California}
\fi 
\if \foreaamo 1
\title{Reduced-Rank Multi-objective Policy Learning and Optimization}
\author{Anonymous Authors}
\fi

\maketitle










\begin{abstract}
Evaluating the causal impacts of possible interventions is crucial for informing decision-making, especially towards improving access to opportunity. However, if causal effects are heterogeneous and predictable from covariates, personalized treatment decisions can improve individual outcomes and contribute to both efficiency and equity. In practice, however, causal researchers do not have a single outcome in mind a priori and often collect multiple outcomes of interest that are noisy estimates of the true target of interest. For example, in government-assisted social benefit programs, policymakers collect many outcomes to understand the multidimensional nature of poverty. The ultimate goal is to learn an optimal treatment policy that in some sense maximizes multiple outcomes simultaneously. To address such issues, we present a data-driven dimensionality-reduction methodology for multiple outcomes in the context of optimal policy learning with multiple objectives. We learn a low-dimensional representation of the true outcome from the observed outcomes using reduced rank regression. We develop a suite of estimates that use the model to denoise observed outcomes, including commonly-used index weightings. These methods improve estimation error in policy evaluation and optimization, including on a case study of real-world cash transfer and social intervention data. Reducing the variance of noisy social outcomes can improve the performance of algorithmic allocations.
\end{abstract}

\section{Introduction} 
Causal inference aims to describe the impact of treatments 
both on aggregate outcomes and also on the trajectories of individuals. When causal treatment effects are heterogeneous, differing systematically across individuals with different covariates, the tailoring of decisions to give treatment mainly to those who are most likely to benefit improves overall welfare for the population. The general problem of \emph{policy learning} is to identify optimal treatment policies using only observed data collected under varying treatment assignments \citep{athey2020pl}. 

Many applications of policy learning arise in social, commercial, or medical domains, where the individuals in question are \emph{people}. Such human-related decision-making poses three fundamental challenges: (1) complex social phenomena are hard to define precisely and consequently researchers generally measure many different outcomes, (2) variability among humans makes outcomes hard to predict, and (3) choosing to measure many outcomes can reduce actionability for optimal decisions by requiring decision makers to choose or specify scalarization weights on the outcomes. 
For example, a recent high-profile ``common task framework" study, the Fragile Families Challenge, 
assessed the ability of machine learning to predict important life outcomes \citep{salganik2020measuring}. 
Though an army of different research teams brought state-of-the-art machine learning models to bear on a large dataset of features and outcomes, the resulting predictions were neither very accurate nor better than a simple baseline model using four features chosen by domain experts. Beyond the caution that this specific study suggests in predicting social outcomes, it highlights the fact that human-related data often involves noisy measurements, ill-defined latent constructs, and multiple outcomes. 

Our focus in the current paper is the problem of multiple outcomes.
The problem is ubiquitous.  In social benefits programs, 
a motivating example for our work that we will illustrate in a case study, poverty is considered to be a dynamic and multidimensional social issue \citep{sen1997poverty}. When attempting to allocate government resources to those in need, measured outcomes may only capture
subsets of dimensions of the true target. Indeed, there are a range of economic variables and social/emotional factors that contribute to the experience of poverty \citep{bossuroy2022poverty, banerjee2022multifacetedpov, blattman2015uganda}. Classically, researchers create an ``index'' or ``proxy'' for these complex measures, which can be as simple as averaging the outcomes in an expert-designated ``category.''
Such approaches are, however, not sensitive to the specific kinds of variables being measured and may lose information by ignoring correlation across outcomes. Several works explore the use of dimensionality reduction \citep{mckenzie2005pcainequality} or otherwise highlight the specific challenges with multiple outcomes \citep{ludwig2017machine,bjorkegren2022machine}.
The problem is also broadly relevant beyond social domains: in e-commerce pricing, search, or recommendation platforms, learning systems measure user behavior as assayed not only via click streams but also, increasingly, a wide variety of forms of ``implicit feedback" that may inform a general propensity to purchase
\citep{garrard2023practical, yao2022efficient, tripuraneni2023choosing}. In healthcare, doctors look at multiple outcomes of interest, including clinical improvement and potential side effects 
to provide individualized treatments to patients optimizing overall health \citep{bastani2021proxies, chen2021itrlatent}. 

In the context of multi-objective policy learning with multiple objectives, the goal is to learn an optimal treatment assignment policy (e.g., subsidy benefit, cash transfer, or financial coaching) that improves multiple outcomes simultaneously, via some weighted average. Within this general problem formulation, our main contributions are as follows. First, we propose an approach to denoising outcome data based on a latent variable model.  Second, we show how to estimate latent outcomes via regression-based optimal control variates for a standard inverse propensity weighting (IPW) estimator.
We show that when learning the optimal policy, denoising reduces variance of estimated values and improves the final policy. 
Finally, we implement our approach with policy optimization for synthetic and real data. We find that denoising outcomes can improve policy evaluation and optimization. Overall, our methods improve actionable decision- and policy-making from noisy outcomes. We show how denoising multiple outcomes, in high-variance social settings, can improve estimation variance, which improves out-of-sample performance of data-driven algorithmic allocations. Our method is able to equitably improve optimal allocation policies derived from noisy, high-dimensional measurements.



\section{Problem Setup and Related Work}

We first summarize our setup. Our data (randomized or observational) consists of tuples of random variables, $\left\{\left(X_i, T_i, Y_i\right): i=1, \ldots, n\right\}$, with $p$-dimensional covariates $X_i \in \mathbb{R}^p$, assigned binary treatment $T_i \in\{0, 1\}$, and vector-valued potential outcomes $Y_i \in \mathbb{R}^k$. (The model is easily generalized to multiple treatments, but we start with binary treatments for brevity). The Neyman-Rubin potential outcomes of applying each treatment option, respectively, are $Y_i(0),Y_i(1)$, and we assume consistency, $Y_i=Y_i\left(T_i\right),$ so that the observed outcome corresponds to the potential outcome of the observed treatment. We assume causal identification, in particular we make the ignorability assumption, $Y(T) \perp T \mid X$, that the covariates sufficiently explain treatment assignment. We also make the stable unit treatment value assumption (SUTVA), i.e., that treatments of others do not affect one individual's outcomes. These are standard, but not necessarily innocuous assumptions in causal inference. Ignorability holds by design, for example, in randomized controlled trials.

We let $\mu_t(X) = \E[Y(t)\mid X]$ denote the outcome model and $e_t(X)=P(T=t\mid X)$ be the propensity score. The policy $\pi(t\mid X)\in[0,1]$ describes the probability of assigning treatment $t$; we seek both to evaluate alternative treatment assignment rules and optimize for the best one. We take the policy $\pi$ to have a restricted functional form, such as a parametrization of a probabilistic classifier, for the purposes of generalization and interpretability. The \textit{policy value estimand} is the population-induced expectation of potential outcomes, if we assigned treatment according to $\pi$. 
Lastly, we consider two relevant settings for multi-objective policy learning: scalarizing observed outcomes $Y(t)$, or, if $Y(t)$ is too high-dimensional, the scalarization of estimated latent factors alone.  We denote the latter as $Z(t)\in\mathbb{R}^r$ where $Y(t)=g(Z(t))$. The scalarized policy value for a weighting vector $\rho \in \mathbb{R}^k$ or $\rho \in\mathbb{R}^r$ is (in the first or second setting, respectively): 
\begin{equation}
    V_{Y}(\pi) = \E[\rho^\top Y(\pi)] \text{ or }     V_{Z}(\pi) = \E[\rho^\top Z(\pi)], 
\end{equation} 
where $\E[\rho^\top Y(\pi)] =\sum_t  \E[ \pi(t\mid X) \rho^\top \E[Y(t)\mid X]]$ and similarly for $\E[\rho^\top Z(\pi)]$.
Unless otherwise stated, we use the convention that the outcomes $Y_i$ are losses so that lower outcomes are better.

\subsection{Estimation in causal inference}
We highlight relevant research on causal inference estimators. General approaches to off-policy evaluation and optimization include the use of the direct method, inverse propensity-weighted estimators, and doubly robust estimators to account for counterfactual outcomes. 

\textbf{Direct Method (DM).} The direct method estimates the relationship between the outcomes conditional on the covariates $X$ and the treatment $T$.   In particular, we use outcome regression to learn $\hat \mu_t(X) = \E\left[Y(t)|X\right]$ for $t=0,1$. 

One standard estimator is based on regression estimation: 
\begin{equation}
\textstyle 
\hat V_{Y}^{DM}(\pi) = \sum_t \E_n\left[ \pi(t\mid X) \mu_t(X)  \right], \label{eqn-dm}
\end{equation}
where we use a logistic policy parametrization, $\pi(t \mid X) = sigmoid(\theta^\top X)$.

\textbf{Inverse Propensity Weighting (IPW).} When $\hat \mu_t(X)$ is a misspecified regression model the resulting estimation procedure can be biased. Under inverse propensity weighting, the observed outcomes are re-weighted by the inverse of the propensity weights, or $\hat \mu_t(X)$ is re-weighted by the inverse of the propensity weights. While IPW is unbiased, dividing by the propensity score can cause the variance to increase. 

The standard IPW estimator is: 
\begin{equation}
\textstyle 
\hat V_Y^{IPW}(\pi) = \sum_t \E_n\left[ \pi(t\mid X) \frac{\indic{ T=t} Y}{e_t(X)} \right]. \label{eqn-ipw}
\end{equation}
\textbf{Doubly Robust (DR).} The doubly robust estimator combines the DM and the IPW estimator \citep{robins2005doublyrobust,scharfstein1999doublyrobust}. 

The standard DR estimator is: 
\begin{equation}
\textstyle 
\hat{V}_Y^{DR}(\pi) = \sum_t \E_n\left[ \pi(t\mid X) \left[\frac{\indic{ T=t}(Y - \mu_t(X)) }{e_t(X)} + \mu_t(X)\right] \right], \label{eqn-dr}
\end{equation}
which has the property of yielding unbiased estimates even if only one of the propensity score model or the outcome model are correctly specified.

\subsection{Reduced Rank Regression for Potential Outcomes}


\textbf{Dimensionality-reducing factor modeling}. Reduced rank regression is a dimension-reducing estimator that corresponds naturally to our setting, via the ``multiple causes to the multiple indicators'' (MIMIC) interpretation \citep{reinsel1998rrr}.  It particular, as we will see, it aligns well with the case studies that motivate our work. Reduced-rank regression (RRR) model structures are often interpreted in terms of an underlying hypothetical construct in which the hypothetical constructs are unobserved ``latent'' variables and there are observed variables that relate the latent variables via a linear model. The general reduced rank regression model can be expressed as follows:
\begin{equation*}
    Y_i = CX_i + \epsilon_i, \hspace{10pt} i=1,\hdots, n,
\end{equation*}
where $Y_i = (y_{1i}, \hdots, y_{ki})$ is a $k\times 1$ vector of outcome variables, $X_i = (x_{1i}, \hdots , x_{pi})$ is a $p \times 1$ vector of covariates, $C$ is the $p \times k$ matrix of regression coefficients, and $\epsilon_i = (\epsilon_{1i},\hdots,\epsilon_{ki})$ is a $k\times 1$ vector of the error terms.

An important assumption of reduced rank regression is that the coefficient matrix $C$ is of lower rank than the original data matrix.
\begin{assumption}[Low rank regression coefficients] 

$rank(C) = r \leq min(k,p)$. 

\label{assumption:lowrank}
\end{assumption}

Under Assumption \ref{assumption:lowrank}, $C$ is the reduced-rank product of two lower-dimensional matrices. Thus, we can write 
\[ C = AB,\]
where $A$ is a $k\times r$ matrix and $B$ is a $r\times p$ matrix, both of which have rank $r$. The $r$ columns of $A$ span the column space of $C$ and the $r$ rows of $B$ span the row space of $C$. The original model can be rewritten as 
\begin{equation*}
    Y_i = A(BX_i) + \epsilon_i, \hspace{10pt} i=1,\hdots, n.
\end{equation*}
This can also be interpreted as a factor model with $Z \coloneqq BX$. \citet{reinsel1998rrr} derive the maximum likelihood estimates $\hat A$ and $\hat B$ for the reduced rank model: 
\begin{equation}
    \hat  A = \Gamma^{-1/2}[\hat V_1,\hdots, \hat V_r], \hspace{10pt} \hat B = [\hat V_1, \hdots, \hat V_r]^\top \Gamma^{1/2}\hat \Sigma_{yx} \hat \Sigma^{-1}_{xx}, \label{eq-rrests}
\end{equation}
where $\hat \Sigma_{yx}  = \frac{1}{n}YX^\top$, $\hat \Sigma_{xx} = \frac{1}{n}XX^\top$, and $V_j$ is the eigenvector corresponding to the j-th largest eigenvalue $\hat \lambda^2_j$ of $\Gamma^{1/2}\hat\Sigma_{yx}\Sigma_{xx}^{-1}\hat\Sigma_{xy}\Gamma^{-1/2}$ and we choose $\Gamma = \hat \Sigma_{\epsilon \epsilon}^{-1}$. 
They show that these estimators are optimal and asymptotically efficient under standard assumptions of independent and normally distributed error terms.  For a discussion of the relationship of this model to the truncated singular value decomposition, see \citet{bunea2011rrroptimal}.

\textbf{Reduced rank regression for potential outcomes}.
In applying this model to our causal setting, we posit RRR models separately for both the treatment and control groups, such that observed outcomes $Y$ are noisy observations.

\begin{assumption}[Multivariate normal $X$]\label{asn-mvnX}
    For  $t=1,0$, we assume that the covariates are $X \sim N(M,\Sigma_x)$. 
\end{assumption}

\begin{assumption}[Potential outcomes satisfy reduced-rank regression assumptions]\label{asn:two-outcome-model}
\begin{align*}
    Z(t) &= B_tX + U \\
    Y(t) &= A_tZ(t) + \epsilon = A_t(B_tX) + (\epsilon + A_tU).
\end{align*} 
\end{assumption}

We estimate the latent and observed outcomes via plug-in estimators: 
\begin{align}&\hat{Z}_1 \coloneqq \hat{B}_1 X,
\qquad \hat{Z}_0 \coloneqq \hat{B}_0 X. \label{eq:Zestimate} \\
&\mu^{RR}_1(X) \coloneqq \hat A_1 \hat B_1 X, \qquad 
\mu^{RR}_0(X) \coloneqq \hat A_0 \hat B_0 X.
\label{eq:YRRRestimate}
\end{align}

The low rank factor model in \Cref{assumption:lowrank} is inspired by the nature of the application problems that we consider (e.g. social data). Additionally, there are classic results in the literature that back up our use of \Cref{asn-mvnX} and \Cref{asn:two-outcome-model} even when the data is skewed or nonlinear. The classical result of \cite{lin2013ols} shows that the coefficient estimate of OLS with full treatment and covariate interactions is asymptotically distributionally consistent and can well approximate more complex data scenarios. Besides, it is common practice in applied fields to report index variables that themselves sum over a category of multiple outcomes, corresponding to a data-agnostic weighting of those outcomes. Our assumption simply formally posits a structure that justifies this common approach and allows for data-driven improvement.


\subsection{Multi-objective evaluation and learning}
In the presence of vector-valued outcomes, as arises in multi-objective optimization, we solve policy optimization problems by scalarizing the multiple outcomes. 

We assume that the decision-maker has already specified a $\rho \in \mathbb{R}^k$ weighting on the original outcomes $Y$.
Then, the optimal policy minimizes the scalarized outcomes:
\begin{equation*}
    \pi^* \in \arg\min \E[ \rho^\top Y(\pi)].
\end{equation*}
An example would be index variables that appear in randomized controlled trials (RCTs) in development economics. Each noisy measured outcome is in a category, such as ``financial well-being" or ``educational progress" and the index simply averages outcomes within a category. We can represent this in our model via restrictions on the $\rho$ vector.\footnote{Namely, consider $\rho$ vectors where every outcome in an index has the same coefficient. Let $\mathcal{I}$ be the set of indices and $I(j) \colon [k] \mapsto [\vert \mathcal{I}]$ describes the index of a variable $Y_j$. Then this common practice corresponds to $\{  \rho \in \mathbb{R}^k \colon \rho_{j} = \rho_{j'} \text{ if } I(j) = I(j'), \;\; \forall j, j' \in [k] \} $.}. Our methods then reduce variance. We focus on this setting for benchmarking our methods, because we can compare to a feasible approach using observed $Y$ outcomes. Our specification of the preference vector in the model is meant to emulate common practices used in economic literature, where practitioners expertly chose an index variable that weights multiple important outcomes of interest.


\textbf{Related work in multi-objective policy learning.} 
Some methodologically different work highlights the ubiquity of multiple outcomes. \citet{viviano2021should} and \citet{ludwig2017machine} study multiple outcomes, but with a focus on hypothesis testing. Recent work investigates short-term surrogates or proxies vs. long-term outcomes \citep{athey2019surrgindex, knox2022socialproxies}, but this clearly designates one as a noisy version of the unavailable other. Instead, we focus on the policy learning setting with multiple outcomes in general, without knowing which one is a proxy for the other. \cite{luckett2021compositeoutcomes} and \cite{lizotte2012reward} both highlight methods that use expert-derived combined outcomes but say that these methods do not account for differences across units. Additionally, a related paper of \cite{bjorkegren2022machine} leverages the many outcomes in an anti-poverty program and instead aims to uncover and audit what policy-makers value in allocation decisions. The approach we take in this paper is complementary and both have the goal of informing better future designs of such policies. 

Some works consider multi-objective policy learning via scalarization, but without denoising. \citet{boominathan2020multiobjPL} study fully observed outcomes. \citet{kennedy2019estimating} consider effect estimation with mean-variance rescaling, but not optimal policies or further denoising. We focus on the benefits of dimensionality reduction for multiple outcomes. Other work uses factor models for policy learning with noisy outcomes.
\citet{chen2021itrlatent} learn discrete-valued latent factors at baseline from a restricted Boltzmann machine model. \citet{saito2022offpolembeddings} consider (known) action embeddings for variance reduction. Our approach differs from previous work 
in estimating continuous constructs in the outcome space.

Factor models and low-rank matrix completion models have been of recent interest in the literature on synthetic control. Our setting is somewhat different: synthetic control requires specific unit-time factor structure, which we don't assume here. We only posit a factor model on the multiple outcomes. We also assume unconfoundedness, whereas in synthetic control the joint unit-time structure and factor model assumption allow for some validity under unobserved confounders. In contrast to synthetic control, we focus on variance reduction for decision-making and we make an observable factor model assumption on outcomes, though assuming unconfoundedness remains uncheckable. 
There are other factor models that we could use besides reduced rank regression in the method of this paper: we discuss some tradeoffs. Another generative factor model is probabilistic PCA \citep{tipping1999probpca}, but it implies an isotropic covariance structure that is incompatible with covariate-adjusted outcomes. \cite{mckenzie2005pcainequality} even go as far as to use PCA on asset data to create an index variable as a measure of inequality, but due to the same issue the authors use the standard deviation of the first PC instead. Other approaches are based on matrix generalizations of nuclear-norm regularization \citep{yuan2007factormodel}, but this is computationally expensive and overestimates the rank.

\section{Methodology}

\begin{algorithm}[t!]\label{alg-policy-learning}
\caption{noise-reduced RR direct method 
}\label{alg-threshlasso}
\begin{algorithmic}[1] 
\STATE{Input: data $(X,T,Y)$, $\rho$, $\Theta$ parameter space for policy $\pi_\theta$
}
\STATE{Standardize the data and outcomes, e.g. demean the data and standardize $X$ to isotropic covariance and divide $Y$ by the standard deviation.}
\STATE{Obtain $\mu_t^{RR}(X)$ by estimating reduced-rank models for $t=0,1$.}
\STATE{Policy learning: Initialize policy $\pi_{\theta}$ 
} 
 \FOR{ t=1,2,$\cdots$}
 \STATE{$\pi_{\theta}^{(t)} = \pi_{\theta}^{(t-1)} - \eta \cdot \nabla_{\theta}V_Y(\pi^{(t-1)}_{\theta})$}
 \ENDFOR
\end{algorithmic}
\end{algorithm}

\subsection{Estimators}

We define a set of estimators based on reduced-rank regression: The first, the direct method, directly imputes counterfactual outcomes via prediction of $Y$ or $Z.$ We also consider variants: denoising observed $Y$ in the inverse-propensity weighted estimator, and a variance-reduced version thereof. 

\textbf{Direct method with reduced rank regression.} We do this using our reduced rank estimation procedure for $\hat A_t \hat Z$. Using reduced rank estimation is a more specialized estimator leveraging the structural assumption of low-rank coefficient matrix. The generic direct method using OLS to estimate each outcome $Y$ yields a full-rank coefficient matrix, and is a baseline that we compare to (and improve upon). 
\begin{equation}\hat V^{RR-DM}_{Y}(\pi) = \sum_t \E_n \left[
\pi(t \mid X) \rho^\top \mu_t^{RR}(X) \right] \label{eqn-est-RRDM}
\end{equation}

\begin{lemma}[Unbiasedness of DM Estimator] 
Under \Cref{asn:two-outcome-model}, the policy value with the direct method estimator (for $\hat Z$ and $\hat Y$) is unbiased:
\[ \textstyle  \E\left[\sum_t \E \left[
\pi(t \mid X) \rho^\top \hat \mu^{RR}_t(X) \right]\right] = V_{Y}(\pi) \] 
\label{lemma-dmunbiased}
\end{lemma}
\textbf{Denoised inverse propensity weighting (IPW).} 
Note that denoised outcomes, $\hat Y,$ could replace observations $Y$ 
in other estimators beyond the direct method. 
We could introduce a variant of the IPW estimator as follows: 
\begin{equation}\hat V^{RR-IPW}_{Y}(\pi) = \sum_t \E_n \left[ \pi(t \mid X)\frac{\indic{T=t}}{e_t(X)} \rho^\top \mu_t^{RR}(X) \right] 
\label{eqn-est-muRR-IPW}
\end{equation}
Under the model, IPW is asymptotically unbiased. 
\begin{lemma}[Unbiasedness of IPW Estimator] Under \Cref{asn:two-outcome-model}, the policy value with the inverse propensity weighting estimator (for $\hat Z$ and $\hat Y$) is unbiased: 
\[\textstyle \E\left[\sum_t\E\left[\pi(t\mid X)\frac{\mathbb{I}[T=t] \rho^\top\mu^{RR}_t(X)}{e_t(X)}\right]\right] = V_{Y}(\pi). 
\]
\label{lemma-ipwunbiased}
\end{lemma}

\subsection{Deriving a control variate estimator}
However, one concern with previous estimators could the reliance on the outcome model. Our last variants add control variates to an IPW or denoised IPW estimator and can have weaker reliance on the model assumption. 

\textbf{Variance reduction via control variates.} 
One benefit of the standard IPW estimator (\cref{eqn-ipw}) is unbiasedness/model robustness, but inverse propensity weighting is also high-variance. Hence variance-reduced estimators via control variates are common in the off-policy evaluation literature. 
The general idea is to add certain zero-mean terms, which can reduce variance due to correlation/anti-correlation with the randomness being averaged in the original estimator. However, when the control variates are vector-valued, a natural question is, what is the optimal linear combination of control variates to minimize variance? We introduce the weighting vector $D_t$ and note that all weighted control variate vectors are also zero mean.

\begin{definition}[Outcome Control Variates]  
For any multivariate function $h_t(X)$, the control variate vector 
is: $$\textstyle C_t(h,X)=\left(1-\frac{\indic{T=t}}{e_t(X)} \right) h_t(X),$$
and for some weighting vector $D_t$, the control variate and variance-reduced IPW estimator are 
\begin{align}
& \phi_t^Y (D_t;\pi) = 
\pi(t\mid X)\left\{\frac{\indic{T=t}\rho^\top Y }{e_t(X)} + D_tC_t\right\}\\
&\hat V^{RR-CV}_{Y} = \sum_t \E_n[ \phi_t^Y (D_t; \pi)].   \nonumber\label{eq-est-cv}
\end{align}

\end{definition}
Above, we add control variates to IPW with the observed outcomes $\rho^\top Y$ for model robustness. But we can also use denoised IPW for further variance reduction, with scalarized $\rho^\top \hat \mu$ or $\rho^\top \hat Z$. We obtain a vector of control variates by choosing $h_t(X) = \hat B_t X$, since $C$ needs to be full rank in the low-dimensional space (i.e. rank = r). 
Different $h_t(X)$ do not change results very much (see appendix) so we proceed with $h_t(X) = \hat B_tX$ for policy optimization.

Finally, \textit{regression control variates} \citep{glynn2002cv} find a weight vector $D$ that achieves optimal variance reduction by maximizing correlation with the randomness in the original estimate. Such a weighting vector is: 
\begin{equation}
     D_t^* = (C_t^\top C_t)^{-1} C_t^\top (\rho^\top  Y)
    \label{eqn-regression-cv}
\end{equation} 
We have that $ D_t$ is the pseudoinverse solution to $\rho^\top  Y = C_t  D_t$. It corresponds to the regression control variate. For our estimated control variate variants, we introduce the estimated weighting vector 
\begin{equation}
     \hat D_t = (C_t^\top C_t)^{-1} C_t^\top (\rho^\top  \hat\mu ).
    \label{eqn-regression-cv-denoised}
\end{equation} 
Although we use \textit{estimates} of $Y$ or $Z$, we show consistency for the regression control variate. 
The next results show that the feasible
variants
also enjoy variance reduction properties (we state it just for $Y$). 
\begin{lemma}[Unbiasedness of CV Estimator]\label{lemma-cvunbiased} Under \Cref{asn:two-outcome-model}, the policy value with the control variate estimator (for $\hat Z$ and $\hat Y$) is unbiased: 
\[ 
 \E\left[\sum_t \E_n[\pi(t\mid X) \phi_t^Y (D_t; \pi)]\right]
= V_{Y}(\pi). \]
\end{lemma}
\begin{proposition}[Consistency in OLS with Noisy Outcomes] Define the oracle weighting vector ${D_t*\coloneqq(C_t^\top C_t)^{-1} C_t^\top (\rho^\top  Y_t)}$. For $t\in \{0,1\}$, \begin{equation}\hat D_t \xrightarrow{p} D_t^*.\end{equation}
\label{prop1:consistency}
\end{proposition}
We abbreviate $\phi_t (D_t)=\phi_t^Z (D_t;\pi)$ when it's clear from context; the following results hold for $\phi_t^Z (D_t;\pi)$ or $\phi_t^Y (D_t;\pi)$. 
\begin{theorem} Assume that $\E\left[CC^\top\right]$ is nonsingular and $\E\left[\rho^\top \hat Z \hat Z^\top \rho + C^\top C\right] < \infty$. Suppose $\E\left[C\right] = 0$ and that $\hat D_t \Rightarrow D^*_t$ as $n \to \infty$. 
Then, for $t \in \{0,1\}$, as $n \to \infty$,
\[ n^{1/2}(\phi(\hat D_t) - \phi( D_t^*)) \Rightarrow 0,\]
so that as $n \to \infty$,
\[ n^{1/2}(\phi(\hat D_t) - \phi) \Rightarrow  N(0,\mathrm{Var}[\phi( D^*_t)]).\]
And as $n \to \infty$, $\phi(\hat D_t)^2 \Rightarrow \E [\phi( D^*_t)^2].$ 
\label{theorem1:asymp-norm}
\end{theorem}

\Cref{theorem1:asymp-norm} shows that our denoised variants of control variate estimator converge to the analogous variance-optimal estimator. It asserts that our control variate estimate $\hat D$ is consistent for $D^*$. 

To summarize, the control variate estimator is:
\begin{align} 
&\hat V_{Y}^{RR-CV}(\pi) = \textstyle
\sum_{t}\E_n[ 
\pi_t(X)\{\frac{\indic{T=t}\rho^\top Y 
}{e_t(X)} + \hat D_tC_t\}
] 
, 
\label{eqn:CV-est}\\ \textstyle 
&C_t\textstyle =\left(1-\frac{\indic{T=t}}{e_t(X)} \right)\hat{B}_t X,   \;\;   D_t = (C_t^\top C_t)^{-1} C_t^\top (\rho^\top Y
).
    \nonumber
\end{align}
Variants replace $\rho^\top Y$ with denoised $\hat Z$ or $\hat Y$. 
\subsection{Analysis }
We can show finite-sample generalization bounds for the out-of-sampling policy value. We assume the policy class is finite, i.e. $\vert \Pi \vert = N$, although this can easily be generalized to infinite policy classes with standard use of covering numbers. We make some standard assumptions regarding bounded outcomes and propensity scores.

 \begin{assumption}[Bounded outcomes]\label{asn-boundedoutcomes} For any $X, T$, the outcomes $|Y_i| \leq L$ almost surely. 
 \label{assumption:boundedoutcomes}
 \end{assumption}
 \begin{assumption}[Overlap in propensity scores]\label{asn-overlap}  For any $x,t$, we have that $0<e_t(x)<1$. 
 \label{assumption:overlap}
 \end{assumption}
 \begin{theorem} Under \Cref{asn-boundedoutcomes} and \Cref{asn-overlap}, and with probability $1-\delta$, we have that
     \begin{align*} &V_{Y}(\pi^*) - V_{Y}^{RR-CV}(\hat \pi_n) \leq 
\textstyle \sqrt{2 \mathrm{log}\left(\frac{2N}{\delta}\right) \cdot \frac{\sup_{\pi\in\Pi}\mathrm{Var}[\phi(D^*; \pi)]}{n}} + \frac{2L}{3n}\mathrm{log}\left(\frac{2N}{\delta}\right) + O_p(n^{-\frac 12})
\end{align*}
 \end{theorem}

The generalization bound depends on the (worst-case) variance of the estimator and therefore illustrates how control-variate estimators improve generalization. The second term estimator-independent depends on the prediction error of estimating $\hat Z$. See appendix analogous results for $\hat Y$. 

\section{Experiments} 
We present experiments on simulated data, and also on a real-world randomized controlled trial for cash transfers with many outcomes.

\subsection{Simulated data}
We demonstrate the performance of our estimators (introducing additional ablations and baselines) 
on simulated data with known ground-truth. The data generation process for these experiments follows the reduced rank latent variable model described in \Cref{asn:two-outcome-model}. We generate the dataset by sampling $n=100$ samples from the following data-generating process. We generate $p$-dimensional $X_i \overset{iid}{\sim} N(M_p,I_p)$ for $i=1,...,n$, where $M_p$ is the mean matrix that is randomly drawn for each dataset. Given ${X}$, the treatment assignment $T_i$ is a Bernoulli random variable, with a logistic treatment probability $P(T_i=1|X_i=x) = e^{\beta^\top x}/(1+e^{\beta^\top x})$. Next we generate 
$B_t$, a $p \times r$ matrix, and $A_t$, a $k \times r$ matrix with independent, standard normally-distributed entries. We set $r=2,k=5,p=8$. We run reduced rank regression on the observed outcomes $Y$ to get estimators $\hat A_t, \hat B_t$.


\subsection{Latent Outcome Estimation Reduces Variance in Off-Policy Evaluation}

Overall, our comparisons illustrate the benefit of denoising noisy outcomes $Y$. We compare the variance reduction of our estimators: direct, IPW, and DR/CV. We report the variance reduction achieved from estimating $Z$ and using control variates. Since practitioners can't compute estimators based on true latent outcomes $Z$, we focus on improvements of the feasible estimator using estimated latent outcomes, $\rho \hat A_t \hat Z_t$. In our experiments, we follow Setting 1. The hypothetical situation is that a decision-maker can specify a weighting/trade-off vector $\rho \in\mathbb{R}^k$: our methods can denoise using predictions of $\hat Y$, and we can compare the improvement of our method vs. feasible estimators that use the original noisy $Y$ outcomes. In the following experiments, we set $\rho = [  0.3987,  0.0212, -0.6195,  1.3661, -1.593 ]$. 


\textbf{Baselines and ablations}. To thoroughly compare the performance of our methods, we introduce additional ablations and baselines, first for dimensionality reduction: 
\begin{itemize}

\item $\rho^\top \mu$ the direct method with OLS (full-rank regression coefficient matrix), 

\item $\rho^\top \mu$-IPW which is IPW replacing $\rho^\top Y$ with $\rho^\top \mu$ (OLS; full-rank coefficients),

\item and $\rho^\top \mu$-CV which is our control variate estimator with $Y$-estimation $\mu$ with OLS rather than our RRR estimates. 

\end{itemize}

\noindent Next we consider baselines with the original $Y$ observations: 

\begin{itemize}

\item $\rho^\top Y$-IPW is standard IPW (\cref{eqn-ipw}),

\item $\rho^\top Y$-DR is the doubly-robust estimator with OLS (\cref{eqn-dr}). 

\end{itemize}

\noindent Next we introduce our proposed methods: 

\begin{itemize}

\item $\rho^\top \hat A \hat Z$ is the direct method with RRR (\cref{eqn-est-RRDM}), 

\item $\rho^\top \hat A \hat Z$-IPW replaces observed outcomes with denoised  (\cref{eqn-est-muRR-IPW}),

\item and $\rho^\top \hat A \hat Z$-CV is our derived control variate estimator (\cref{eqn-regression-cv-denoised}). 
\end{itemize}

\begin{figure}[t!]
\centering
\includegraphics[width=0.45\columnwidth]{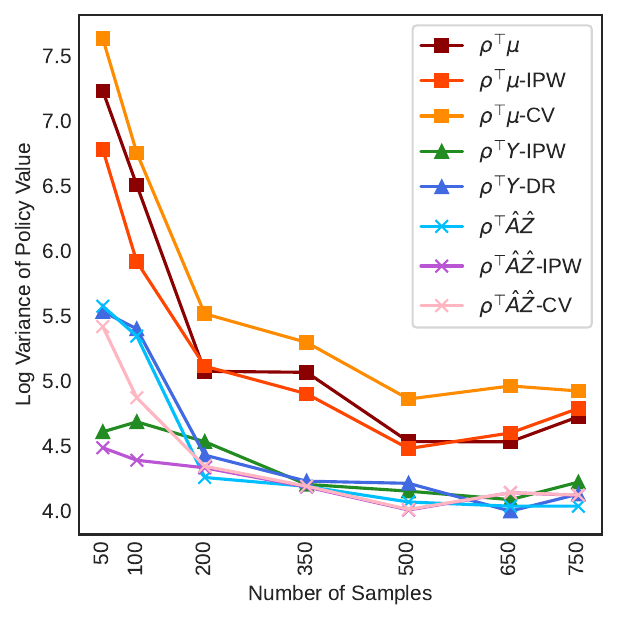}
\caption{ Policy evaluation experiment: comparing variances of the policy value for each estimator averaged over 100 datasets. We compute each policy value by using the optimal policies learned under the true latent outcomes $\rho^\top A_tZ$. \textit{Lower is better.}}
\label{fig:policy_eval}
\end{figure}

\begin{figure}[t!]
\centering
\includegraphics[width=0.35\columnwidth]{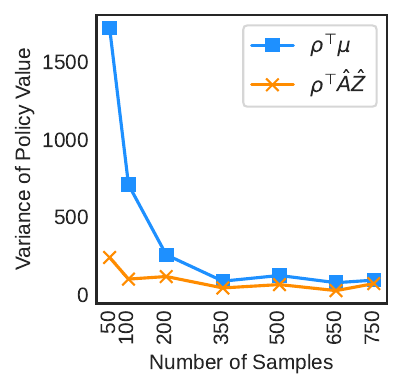}
\includegraphics[width=0.33\columnwidth]{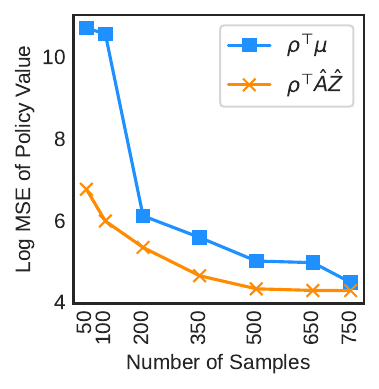}
\label{fig:po_directmethod}
\includegraphics[width=0.35\columnwidth]{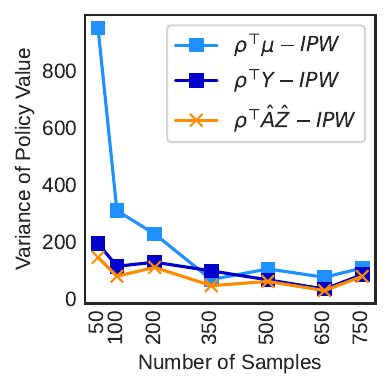}
\includegraphics[width=0.33\columnwidth]{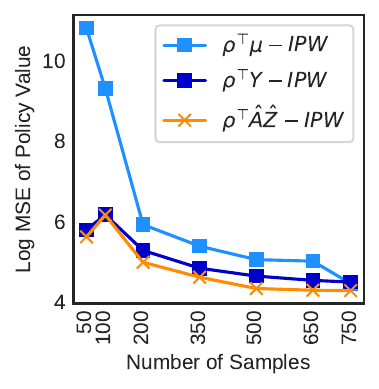}
\label{fig:po_ipwmethod}
\includegraphics[width=0.35\columnwidth]{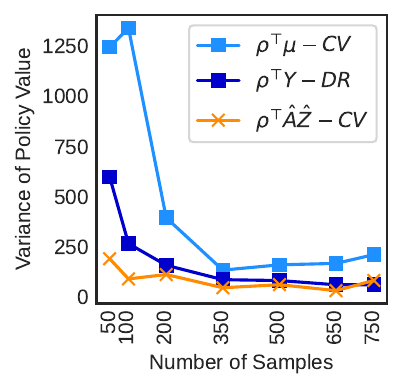}
\includegraphics[width=0.33\columnwidth]{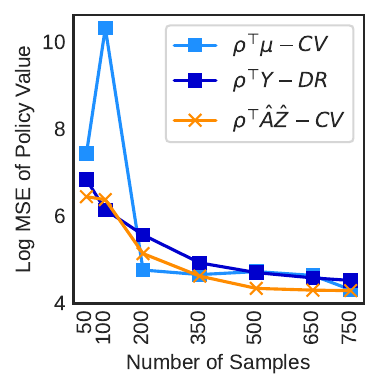}
\caption{ Policy optimization experiment: left figures illustrate variance of out-of-sample policy value estimate (averaged over 50 datasets). Right figures compare the log MSE for policy value suboptimality.
\textit{Lower is better.} Top row compares $\rho^\top \hat A \hat Z$ (reduced rank DM) with full-rank DM. Middle row compares standard IPW with our denoised IPW
, and bottom row compares standard doubly robust estimator ($\rho^\top Y$-DR) with our control variate estimator.
}\label{fig:po_cvmethod}
\end{figure}

Later in \Cref{fig:policy_eval} we compare all of these methods and in \Cref{fig:po_directmethod} we compare each estimator variant to its most relevant baselines and ablations. 
\begin{figure}[t!]
\centering
\includegraphics[width=0.33\columnwidth]{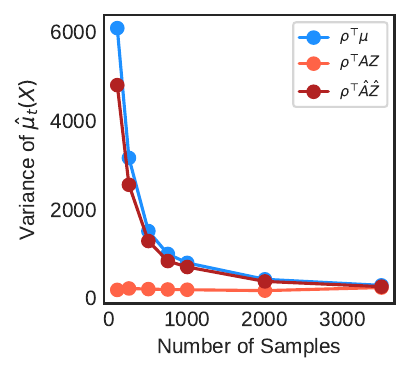}
\includegraphics[width=0.35\columnwidth]{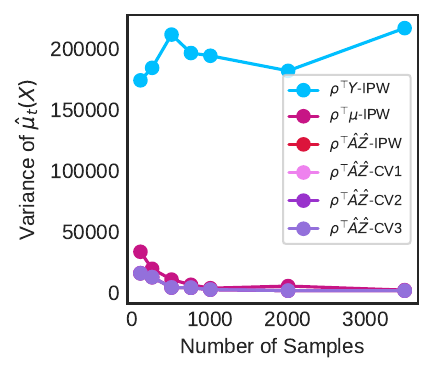}
\caption{Variance in ATE estimation. \textbf{Left:} Comparison of variances averaged over 100 datasets as the sample size of the dataset increases using the direct estimators for $\rho Y$. \textbf{Right:} Comparison of variances as the sample size increases for the IPW and control variate estimators. \textit{Lower is better.}}
\label{fig:est_variance}
\end{figure}
\Cref{fig:est_variance} compares the variance of all the estimators for the ATE as the sample size increases. We observe significant variance reduction due to denoising $Y$, which somewhat overpowers the added benefit of using the control variates. In the appendix, we include additional experiments in \Cref{fig:vary_dimensions,fig:varying_noise} to assess how changing relative dimension of $Y$ and $Z$ and the observation noise level of $Y$ affects variance reduction. 

Our figures show that the ablations $\rho^\top \mu, \rho^\top \mu-$IPW, $\rho^\top \mu$-CV generally perform worse than their reduced-rank regression counterparts, indicating that it is our structural assumption of lower-dimensional constructs that leads to significant variance reduction rather than our parametric linear model specification.

\textbf{Policy evaluation.} For policy evaluation, in \Cref{fig:policy_eval} we compare all of the estimators evaluated on an estimate of the oracle-optimal policy $\pi^*$ (i.e., obtained by running policy optimization on the true latent outcomes $\rho^\top A_t Z$). We evaluate the variance reduction achieved in the objective value and policy value and use the optimal policy $\pi^*$. To obtain $\pi^*$, we run the optimization procedure with subgradient descent for 20 iterations. We  evaluate the estimators over 100 trials. In \Cref{fig:policy_eval}, $\rho^\top \hat A_t \hat Z$ improves especially when the sample size is small. We see the benefits of denoising when compared to our ablations and even the best competitor, the doubly robust estimator.

Notably, for every estimator, the denoised version does better. We recommend standardizing outcomes (i.e. so that the simple average is the precision-weighted average), and recall that IPW can be thought of as a doubly-robust estimator imputing 0 for counterfactual outcomes. Although in the simulated data, we did not standardize outcomes, the outcomes are zero-mean in the data-generating process. In this setting, combining the denoised $\hat Y$ estimate with IPW does well and is similar to the CV-based estimator: we think this is because IPW with standardized outcomes conducts additional shrinkage towards the mean.

\textbf{Evaluating multi-objective policy optimization}. Lastly, in \Cref{fig:po_cvmethod} we run policy optimization with all of the estimators where we use a subgradient method to carry out the policy optimization. We run the optimization procedure over 40 iterations for each of the 50 trials. We get a $\hat \pi$ for each of the estimation methods and compute the policy value on an out-of-sample dataset using $\hat \pi$. We also compute the optimal policy $\pi^*$ and compare it to $\hat \pi$ using the mean squared error.

\begin{table*}[t!]
\caption{
Out-of-sample variance for DM, IPW, DR, and CV estimators under policy evaluation (\textbf{lower} is better).}
\label{policy-eval-table}
\vskip 0.1in
\begin{center}
\begin{small}
\begin{tabular}{lcccccccc}
\toprule
Estimator & $\rho^\top \mu^{DM}$  & $\rho^\top \mu^{RR-DM}$ & $\rho^\top Y^{IPW}$ & $\rho^\top \mu^{IPW}$ & $\rho^\top \mu^{RR-IPW}$ & $\rho^\top Y^{DR}$ & $\rho^\top \mu^{CV}$ &$\rho^\top \mu^{RR-CV}$\\
\midrule
Out-of-Sample Variance & 4.373 & 4.373 & 111.746 &  8.369 & 5.009 & 111.236 & 2.757 & 5.058 \\
\bottomrule
\end{tabular}
\end{small}
\end{center}
\vskip -0.1in
\end{table*}

\begin{table*}[t!]
\caption{Out-of-sample evaluation of optimized policy value (\textbf{higher} is better). Rows indicate policies optimizing different estimates on the training set. Columns indicate different evaluation methods on the test set. Estimated factors $\hat Z$ = 2. We report the mean $\pm$ standard deviation. Bold is best (within-column), italic second-best.}
\label{model-robust-table}
\vskip 0.1in
\begin{center}
\begin{small}
\begin{tabular}{lcccc}
\toprule
$\pi^*$& Self  & $\rho^\top \mu^{RR-DM}$   & $\rho^\top Y^{IPW}$   & $\rho^\top Y^{DR}$  
\\
\midrule
 $\rho^\top \mu^{DM}$ &0.101 $\pm$ 1.836   & -0.126 $\pm$ 1.037         & \textbf{0.180} $\pm$ 11.959  & \textbf{0.177} $\pm$ 11.908  
 
 \\
 $\rho^\top \mu^{RR-DM}$  & -0.006 $\pm$ 0.909  & \textbf{-0.006} $\pm$ 0.909         & \textit{0.144} $\pm$ 11.461  & \textit{0.147} $\pm$ 11.407  
 
 \\
 $\rho^\top Y^{IPW}$ & -0.224 $\pm$ 10.120 & -0.330 $\pm$ 1.026         & -0.224 $\pm$ 10.120 & -0.225 $\pm$ 10.094 

 \\
 $\rho^\top \mu^{RR-IPW}$ & -0.045 $\pm$ 1.227  & \textit{-0.044} $\pm$ 0.919         & 0.046 $\pm$ 11.217  & 0.046 $\pm$ 11.166 

 \\
 $\rho^\top Y^{DR}$ &-0.258 $\pm$ 10.244 & -0.305 $\pm$ 1.029         & -0.253 $\pm$ 10.276 & -0.258 $\pm$ 10.244 

 \\
 $\rho^\top \mu^{RR-CV}$  & -0.045 $\pm$ 1.231  & \textit{-0.044} $\pm$ 0.919         & 0.046 $\pm$ 11.217  & 0.046 $\pm$ 11.166 

 \\
\bottomrule
\end{tabular}
\end{small}
\end{center}
\vskip -0.1in
\end{table*}
\subsection{Real world case study: ``Sahel" dataset, poverty graduation program}

We turn to a dataset derived from a large scale multi-country randomized control trial designed to evaluate social safety net programs aimed at graduating households out of poverty \citep{bossuroy2022tackling}. The program duration spanned 2 to 5 years and the data was collected in four countries including Niger, Senegal, Mauritania, and Burkina Faso. 

\textbf{Features.} The dataset contains baseline characteristics measured at the program start. The survey used to collect this information included questions about productive and non-productive household activities, housing, assets, familial relationships, psychology and mental health, food consumption spending, education spending, and health spending.  
The final dataset used for experimentation consists 4,139 households and 36 features. 

\textbf{Treatments.} Although the original experiment considered multiple treatment levels that are different combinations of six previously-identified financial or psycho-social interventions, we focus on a subset of treatments: control arm (no intervention) and full treatment (received all six interventions). See the appendix for more details. 

\textbf{Outcomes.} Program effects were measured not only with two primary outcomes (food insecurity and consumption per capita
), but also a large set of secondary outcomes. 
We selected \emph{seven} outcomes (direct program targets), screened for treatment effect heterogeneity. These include beneficiary wages, food consumption, business health, beneficiary productive revenue, business asset value, investments, and savings. Note unlike our previous generic assumption on cost outcomes, these are benefits that we would like to \emph{maximize}.   

\textbf{Preprocessing.} 
We split into training (3,311 households, for optimizing policies) and test datasets (828 households, for out-of-sample off-policy evaluation) and standardize (i.e. subtract the mean and divide by the standard deviation) for both outcomes $Y$ and features $X$. 

\textbf{Policy evaluation.} 
We assess variance reduction for policy evaluation. We (arbitrarily) evaluate a near-optimal policy: we obtain $\hat\pi^*_{DM},$ by optimizing a naive direct method estimator with OLS. We set $\rho = [8.28,1.31,0.21,0.061,0.59,0.01,1.12]$.

Given that we already used the training dataset to find a near-optimal policy, we assess the out-of-sample variance on the test set. In \Cref{policy-eval-table}, we see that the variance is much lower when we use our model to estimate $\hat Y$. Although the bias remains unknown in real data, we obtain evidence of overall improvement in the next policy optimization experiment. 

\textbf{Policy optimization.} 
Ultimately we would like to learn a treatment policy that \textit{maximizes} these financial outcomes for each household. We learn optimal policies using the range of methods described previously in ``Baselines and Ablations." 
Since this is a real dataset, we don't have the true counterfactual outcomes for observed individuals, but we can again use off-policy evaluation to \textit{estimate} the value of the learned policies. 
In this real-world data experiment, where we do not know the true underlying data generating process, we ensure best possible estimation of $\hat Z$ by conducting a hyperparameter sweep for our reduced rank regression estimates (see appendix for details). We set $\rho =[8.28,1.31,0.21,0.061,0.59,0.01,1.12]$.

In \Cref{model-robust-table}, we provide the out-of-sample estimates of the policy value using a series of different policies $\hat \pi^*$: each row optimizes a different policy value estimator. The best that we can do is evaluate on the held-out test set. Columns indicate different evaluation methods: first we evaluate each policy against its own estimation method, and then with the model-robust IPW and DR methods. The last two columns provide model free evidence of our $\hat Z$ estimation procedure being robust to model misspecification. Across all the evaluation techniques (table columns), we see consistent improvement in the policy value and the within-column orderings of each evaluation is relatively stable. To summarize, our methods ($\rho^\top \mu^{RRR}$ with DM, IPW, or CV) are consistently better than no denoising at all. Our denoised method has counterintuitive practical benefit in such social data settings, where common folklore is that causal effects in the social sciences are very noisy zeros and many objectives are noisy realizations of some underlying notion. Empirically, we see the variance reduction and improved causal estimation from our proposed method which ultimately leads to improved policy learning. From our learned treatment policies, we see that the following features: household head being female, household head age, household head years of education, and difficulty rating for lifting 10kg bag by household head are the most important features that result in being treated by the policy. 


\section{Conclusion}
We highlighted how multiple outcomes can pose estimation challenges, and developed tools using dimensionality reduction in multivariate regression to reduce variance and improve policy evaluation and optimization. Directions for future work include reducing model dependence via standard reweighted maximum likelihood estimation for the outcome model, or more advanced nonlinear dimensionality reduction, and employing model selection.

\section{Impact Statement}

Our work is methodological in nature but we in particular focus on social settings, so care is warranted. Important considerations include domain-driven discussion about non-discriminatory policies and potential additional concerns about fairness constraints. Latent variables in particular have a history of being used in the social sciences in overly reductive or essentialist ways, which we caution against in general. Our focus here is on variance reduction, perhaps even given already-existing broad categorizations like ``financial well-being" and ``educational outcomes", but we recommend additional use of interpretive tools in practice to interpret and validate latent factors.










\bibliographystyle{plainnat}
\bibliography{ref.bib}

\appendix
\onecolumn

\section{Additional discussion}

\subsection{Related work}
\textbf{Other factor models.} Although we focus on reduced rank regression, we briefly describe
some advantages related to other possible latent factor models, some of which could be equivalently adapted. Another probabilistic latent variable is probabilistic principal component analysis (PPCA) \citep{tipping1999probpca}. A common latent variable model assumes a linear relationship between latent and observed variables; PPCA assumes
that, conditional on the latent variables, the observed data Y are Gaussian.
The marginal distribution over the latent variables is therefore \textit{isotropic} Gaussian, and the implication of equal variance is a poor fit for heterogeneous covariate-adjusted outcomes.
Another option is a penalized least squares factor model. 
\citep{yuan2007factormodel} proposes simultaneous estimation of factors and factor loadings with a sparsity penalty, generalizing ridge regression for factor estimation. Unfortunately they penalize the coefficient matrix's Ky Fan norm, which is computationally intensive and overestimates the rank.

\section{Proofs}
\begin{proof}[Proof of \Cref{lemma-dmunbiased}] 

\begin{align}
    &\E \left[ \sum_t \E \left[ \pi(t\mid X) \rho^\top \hat Z_t \right]\right] - \sum_t \E\left[ \pi(t\mid X) \rho^\top \E\left[ Z(t) \mid X\right]\right] \\ 
    &= \sum_t \E \left[ \pi(t\mid X) \rho^\top \E \left[ \hat B_t X \mid X \right]\right] - \E\left[ \pi(t\mid X) \rho^\top \E\left[ B_t X\mid X\right]\right] \\ 
    &= \sum_t \E \left[ \pi(t\mid X) \rho^\top \E \left[ \hat B_t X - B_t X\mid X\right]\right] \\
    &= \sum_t \E \left[ \pi(t\mid X) \rho^\top \E \left[ (\hat B_t- B_t)X\mid X\right]\right] \\
    &= 0 
\end{align}

where lines (14) and (15) are true by definition of $\hat Z_t$, and under \Cref{asn:two-outcome-model}. In the last line, we used the fixed-design unbiasedness of $\E[(\hat B_t  - B_t)X \mid X] = 0$. Theorem 2.4 from \citet{reinsel2022multivariate} gives the asymptotic normality result for $\hat B_t$ that implies asymptotic unbiasedness. 

The result for $\hat Y $ holds using prediction error unbiasedness, where we have that $\E[(\hat A_t\hat B_t  - A_tB_t)X \mid X] = 0$ by the asymptotic normality results of Equation 2.36 in \citet{reinsel2022multivariate}.

\end{proof}

\begin{proof}[Proof of \Cref{lemma-ipwunbiased}] 

\begin{align}
    &\E \left[ \sum_t \E \left[ \pi(t\mid X) \frac{\indic{T=t}}{e_t(X)}\rho^\top \hat Z_t \right]\right] - \sum_t \E\left[ \pi(t\mid X) E\left[ \frac{\indic{T=t}}{e_t(X)}\rho^\top Z(t) \mid X\right]\right] \nonumber\\ 
    &= \sum_t \E \left[ \pi(t\mid X) \rho^\top \E \left[ \frac{\indic{T=t}}{e_t(X)}\hat Z_t  \mid X \right]\right] - \sum_t \E\left[ \pi(t\mid X) \frac{\E[\indic{T=t}\mid X]}{e_t(X)}\rho^\top \E\left[ Z(t) \mid X\right]\right] \label{eq-pfipw-iterexp} \\
    &= \sum_t \E \left[ \pi(t\mid X) \rho^\top \frac{\E[\indic{T=t}\mid X]}{e_t(X)}\E \left[ 
    \hat B_t X 
    \mid X \right]\right] - \E\left[ \pi(t\mid X) \rho^\top E\left[ Z(t) \mid X\right]\right] 
    \label{eq-pfipw-pullout} 
    \\
    &= \sum_t \E \left[ \pi(t\mid X) \rho^\top \E \left[ \hat B_t X \mid X \right]\right] - \E\left[ \pi(t\mid X) \rho^\top E\left[ B_t X\mid X\right]\right]  \label{eq-pfipw-ipwcancels}\\ 
    &= \sum_t \E \left[ \pi(t\mid X) \rho^\top \E \left[ \hat B_t X - B_t X\mid X\right]\right]  \nonumber\\
    &= \sum_t \E \left[ \pi(t\mid X) \rho^\top \E \left[ (\hat B_t- B_t)X\mid X\right]\right] \nonumber \\
    &= 0  \nonumber
\end{align}

where \cref{eq-pfipw-iterexp} follows by iterated expectations, \cref{eq-pfipw-pullout} applies the pull-out property of conditional expectation, and 
\cref{eq-pfipw-ipwcancels}
is true by definition of the propensity score. The rest follows by the assumption on the data generating model \Cref{asn:two-outcome-model}. In the last line, we again used the fixed-design unbiasedness of $\E[(\hat B_t  - B_t)X \mid X] = 0$. Theorem 2.4 from \citet{reinsel2022multivariate} gives the asymptotic normality results for $\hat B_t$ that implies unbiasedness. 

The result for $\hat Y $ holds using prediction error unbiasedness, where we have that $\E[(\hat A_t\hat B_t  - A_tB_t)X \mid X] = 0$ by the asymptotic normality results of Equation 2.36 in \citet{reinsel2022multivariate}.

\end{proof}

\begin{proof}[Proof of \Cref{lemma-cvunbiased}]
The proof essentially follows by construction. Without loss of generality we omit the $\pi(t\mid X)$ term from the below. The first part of our estimator is the IPW estimator. Assuming the propensity scores are well specified, then we have that 

\[ \textstyle  \E\left[(1 - \frac{\indic{T=1}}{e_1(X)}) \mid  X\right] = 0.\]  

This implies that $\forall h(X)$ functions, 

\[\textstyle 
    \E\bigg[(1-\frac{\indic{T=1}}{e_1(X)})h(X) \bigg] = \E \bigg[\E[(1-\frac{\indic{T=1}}{e_1(X)}) \mid X] h(X)\bigg] = 0
\] 

The first equality is true by iterated expectation. This is mean-zero for all functions of X. We have 

\[ \E \bigg[ \frac{\indic{T=1} \rho^\top \hat Z}{e_1(X)} + \hat D_1 C_1 | X, T\bigg] =\underbrace{\E\bigg[\frac{\indic{T=1}\rho^\top \hat Z}{e_1(X)}\bigg]}_{\text{consistent estimator}} + \underbrace{\E[\hat D_1 C_1]}_{\approx \text{ mean-zero noise}} \]

The proof of the result for $\E \bigg[
\pi(t\mid X)\{ \frac{\indic{T=0} \rho^\top \hat Z}{e_0(X)} + \hat D_0 C_0 \}\bigg]$ is similar. Thus our CV estimator $\hat \phi_t(\hat D_t)$ is unbiased. 

The result for $\hat Y $ follows since the IPW estimator is also unbiased for $\hat Y$ by \cref{lemma-ipwunbiased}.   
\end{proof}

\begin{proof}[Proof of \Cref{prop1:consistency}]
We invoke standard assumptions in measurement noise of the dependent variable in OLS, which says that the measurement error is uncorrelated with $X$. Recalling that $\hat B = V^\top \Gamma^{1/2}\Sigma_{yx}\Sigma^{-1}_{xx}$, let $\epsilon_{\hat Z} = \hat B_t X - B_t X$ denote the measurement error in $\hat Z$; it is the estimation error from reduced rank regression. Then 
\begin{align*} \mathrm{Cov}(\epsilon_{\hat Z}, X) &= \E\left[\epsilon_{\hat Z} X\right] - \E\left[ \epsilon_{\hat Z} \right]\E\left[ X \right] = \E\left[\epsilon_{\hat Z} X\right] \\
&= \E\left[ \E\left[ \epsilon_{\hat Z}\mid X\right] X\right]  = 0 
\end{align*}
where the last inequality follows from $X$-conditional unbiased estimation of reduced rank regression parameters \citep[p. 184]{reinsel2022multivariate}. 

Hence OLS with $\hat Z$ is consistent, in effect as if there were no measurement error \citep[4.4.1]{wooldridge2001measure}. 
 
\end{proof}

\begin{proof}[Proof of \Cref{theorem1:asymp-norm}]
Assuming that $\E\left[C_tC^\top_t\right]$ is nonsingular, then $\hat D^*_t \in \argmin_D \mathrm{Var}\left[\phi(D)\right]$ is given by 

\[ \hat D^*_t = \E\left[C_tC^\top_t\right]^{-1}\E\left[\frac{\indic{T=t}\rho^\top \hat Z_t}{e_t(X)}\right] \hat B_t X_t \] 

and 
\[ \hat D_t = \E_n\left[C_tC^\top_t\right]^{-1}\E_n\left[\frac{\indic{T=t}\rho^\top \hat Z_t}{e_t(X)}\right] \hat B_t X_t\] 

We can invoke the results of Theorem 1 from \citep{glynn2002cv} and \Cref{prop1:consistency} to get asymptotic equivalence for our control variate estimator. 
\end{proof}

\section{Generalization bounds for policy learning }

\subsection{Preliminaries}
First we begin by collecting some technical results used without proof.

\begin{lemma}[Proposition 15 from \citet{bunea2011rrroptimal}] Let $E$ be a $n \times k$ matrix with independent subGaussian entries $E_{ij}$ with subGaussian moment $\Gamma_{E}$. Let $X$ be an $n \times p$ matrix of rank $q$ and let $P=X(X^{\top}X)^{-}X^{\top}$ be the projection matrix on $R[X]$. Then, for each $x>0$, and large enough $C_0$, we have 
\[\Pr(d_1^2(PE) \leq C_0(k+q))\leq 2 \exp{(-(k+q))},\]
where $d_1(PE)$ is the largest singular value of the projected noise matrix $PE$. 
\label{lemma:stochastic-bound}
\end{lemma}

\begin{lemma}[Error bounds for the RRR rank selection criterion estimator (\citet{bunea2011rrroptimal}, Theorem 5)] This bound is derived for the fit $$\| \hat A_t \hat B_t X - A_t B_t X  \|^2_F \coloneqq \sum_i \sum_j\left\{(\hat A_t \hat B_t X  )_{i j}-(A_t B_t X  )_{i j}\right\}^2$$ based on the restricted rank estimator for each value of the rank $r$. Then for large enough constants $C_1, C_2$ and with probability one, we have 
\[\| \hat A^{(r)}_t \hat B^{(r)}_t X - A_t B_t X  \|^2_F \leq C_1 \sum_{j>r}d_j^2(A_t B_t X ) + C_2rd_1^2(PE),\]

where $\sum_{j>r}d_j^2(A_t B_t X )$ is an approximation error and $rd_1^2(PE)$ is a stochastic term that concentrates the error, is increasing in $r$, and can be bounded by a constant times $r(k+q)$ by \cref{lemma:stochastic-bound}.  
\label{lemma:rrr-bound}
\end{lemma}
This bound holds when we choose the rank based on the rank selection criterion described in Section 2 of \citet{bunea2011rrroptimal}. Although we did not choose our rank in this way, we find that empirically the choice of rank does not matter much in our experiments. 







\begin{lemma}[Bernstein's inequality] Suppose $X_1, \hdots, X_n$ are i.i.d. with 0 mean, variance $\sigma^2$ and $\norm{X_i}{} \leq M$ almost surely. Then with probability at least $1-\delta,$ we have \[ |\frac{1}{n}\sum_{i=1}^n X_i| \leq \sqrt{\frac{2\sigma^2}{n}\mathrm{log}\left(\frac{2}{\delta}\right)} + \frac{2M}{3n}\mathrm{log}\left(\frac{2}{\delta}\right).\] \label{lemma-bernsteinineq}   \end{lemma}
    




\subsection{Results}

 \begin{assumption}[Bounded outcomes]For any $X, T$, the outcomes $|Y_i| \leq L$ almost surely. 
\label{assumption:boundedoutcomes}
 \end{assumption}
     
 \begin{assumption}[Overlap in propensity scores]For any $X,T$, $0<e_t(X)<1$. 
 \label{assumption:overlap}
 \end{assumption}

 \begin{theorem}[Deviation bound for CV]\label{thm-deviationbound} Under \Cref{assumption:boundedoutcomes}
 and \Cref{assumption:overlap}, further assume that $\abs{\hat V^{RR-CV}_{\hat Y}(\pi)}{} \leq L \cdot 1$. Then with probability at least $1-\delta$, we have 
\[ \abs{\hat V^{RR-CV}_{\hat Y}(\pi) - V_Y(\pi)}{} \leq \sqrt{2 \mathrm{log}\left(\frac{2}{\delta}\right) \cdot \frac{\mathrm{Var}[\phi^Y(D^*; \pi)]}{n}} + \frac{2L}{3n}\mathrm{log}\left(\frac{2}{\delta}\right) + 
\sqrt{\frac 2n}
\{C_1 \sum_{j>r}d_j^2(A_t B_t X ) + C_2rd_1^2(PE)\}^{\frac 12}. \]
and
\[ \abs{\hat V^{RR-CV}_{\hat Z}(\pi) - V_{Z}(\pi)}{} \leq \sqrt{2 \mathrm{log}\left(\frac{2}{\delta}\right) \cdot \frac{\mathrm{Var}[\phi^Z(D^*; \pi)]}{n}} + \frac{2L}{3n}\mathrm{log}\left(\frac{2}{\delta}\right) + 
\sqrt{\frac 2n}
\{C_1 \sum_{j>r}d_j^2( B_t X ) + C_2rd_1^2(PE)\}^{\frac 12}. \]
\label{theorem:devboundcv}
 \end{theorem}

 \begin{remark}[Variance reduction for regression control variates] The bound above is dependent on the worst-case variance across the policy class. By construction, our regression control variate estimator is one that minimizes variance. To see this, note that the variance of the estimator with control variates is: 

\[ \mathrm{Var}[\phi^{CV}(D^*)] = \mathrm{Var}\bigg[\frac{\indic{T=1}\rho^\top \hat Z}{e_1(X)}\bigg] - 2 D^* \E\bigg[ \frac{\indic{T=1}\rho^\top \hat Z}{e_1(X)}\bigg] C + D^\top \E[CC^\top] D. \]

Note that $D=0$ is a feasible solution for this optimization problem but it is not optimal, therefore our choice of $D^*$ has a lower variance than the IPW estimator.
\[ \mathrm{Var}[\phi^{CV}(D^*)] \leq \mathrm{Var}\bigg[ \frac{I[T=t]\rho^T \hat Z}{e_t(X)}\bigg]\]

The result holds for $\hat Y$ as the proof relies only on our choice of $D^*$. 
\label{remark-cvvariance}
\end{remark} 

\begin{corollary}[Uniform deviation bound for policy optimization] Given a finite policy class $\Pi$ with $|\Pi| = N$, assume that the conditions of \Cref{theorem:devboundcv} are satisfied for each $\pi \in \Pi$. Then with probability at least $1-\delta$, we have 
\[ \sup_{\pi\in\Pi} \abs{\hat V^{RR-CV}_{\hat Y}(\pi) - V_Y(\pi)}{} \leq \sqrt{2 \mathrm{log}\left(\frac{2N}{\delta}\right) \cdot \frac{\mathrm{Var}[\phi^Y(D^*; \pi)]}{n}} + \frac{2L}{3n}\mathrm{log}\left(\frac{2N}{\delta}\right) + 
\sqrt{\frac 2n} 
\{C_1 \sum_{j>r}d_j^2(A_t B_t X ) + C_2rd_1^2(PE)\}^{\frac 12}
.\]
\label{corollary:devboundpopt}
\end{corollary}

\begin{theorem}[Generalization error]\label{theorem:statanalysis} With probability $1-\delta$, we can bound our estimate of $V^{RR-CV}(\hat \pi_n)$ by 
\[ V^{RR-CV}(\hat \pi_n;\hat Y) \geq V_Y(\pi^*) - \sqrt{2 \mathrm{log}\left(\frac{2N}{\delta}\right) \cdot \frac{\sup_{\pi \in \Pi}\{ \mathrm{Var}[\phi^Y(D^*; \pi)] \}}{n}} + \frac{2L}{3n}\mathrm{log}\left(\frac{2N}{\delta}\right) + 
\sqrt{\frac 2n} 
\{C_1 \sum_{j>r}d_j^2(A_t B_t X ) + C_2rd_1^2(PE)\}^{\frac 12}
\]
\end{theorem}

\subsection{Proofs}

\begin{proof}[Proof of \Cref{thm-deviationbound}]
To analyze the finite-sample error, we decompose by 
$\pm \E[ \hat V^{RR-CV}_{ \hat Y}(\pi)]$
to get two terms that represent the generalization error and estimation error, 
\begin{align*}
\abs{\hat V^{RR-CV}_{\hat Y}(\pi) - V_Y(\pi)}{} &\leq 
\abs{  \hat V^{RR-CV}_{\hat Y}(\pi) - V_Y(\pi) \pm  
\E[ \hat V^{RR-CV}_{ \hat Y}(\pi)]}
\\
&\leq \underbrace{  \abs{
\hat V^{RR-CV}_{\hat Y}(\pi) - 
\E[ \hat V^{RR-CV}_{ \hat Y}(\pi)]}
}_{\text{generalization error}}+ \underbrace{\abs{
\E[ \hat V^{RR-CV}_{ \hat Y}(\pi)] -  V_{ Y }(\pi)}{} }_{\text{estimation error}  }
\end{align*}

We start by bounding the first term. By \Cref{lemma-cvunbiased}, we know that $\hat V^{RR-CV}_{\hat Y}(\pi)$ is a sum of $n$ i.i.d. random variables, each with mean $\E[ \hat V^{RR-CV}_{ \hat Y}(\pi)]$. We also have bounds on both the range and variance of the CV estimator from \Cref{lemma-cvunbiased} and \Cref{remark-cvvariance}. We can immediately obtain an upper bound by the application of Bernstein’s inequality in \Cref{lemma-bernsteinineq} uniformly over the policy space 
\[\abs{
\hat V^{RR-CV}_{\hat Y}(\pi) -
\E[ \hat V^{RR-CV}_{ \hat Y}(\pi)]
}{} \leq \sqrt{2 \mathrm{log}\left(\frac{2}{\delta}\right) \cdot \frac{\mathrm{Var}[\phi^Y(D^*; \pi)]}{n}} + \frac{2L}{3n}\mathrm{log}\left(\frac{2}{\delta}\right)\]

Next, we bound the second term as follows
\begin{align*}
 \abs{V_{ Y }(\pi)-\E[ \hat V^{RR-CV}_{ \hat Y}(\pi)] }
 &= \abs{\sum_t \E\left[  \pi(t\mid X) \frac{\indic{T=t}}{e_t(X)} \rho^\top Y \lvert X,T\right] - \sum_t\E\left[ 
  \pi(t\mid X)
  \frac{\indic{T=t}}{e_t(X)} \rho^\top \hat Y \lvert X,T \right]}{} \\ 
 &= \abs{
 \E\left[ \sum_t  \pi(t\mid X)(\rho^\top A_t B_t X- \rho^\top \hat A_t \hat B_t X)
 \right] }{}\\
  &=\abs{
  \E \left[ \pi(0\mid X) \rho^\top(A_0 B_0 - \hat A_0 \hat B_0)X +  \pi(1\mid X)\rho^\top B_1 - \hat A_1 \hat B_1)X \right] 
  }  \\ 
 & \leq \abs{\E[ \pi(0\mid X) \rho^\top(A_0B_0 - \hat A_0 \hat B_0)X ]}{} + \abs{ \E[ \pi(1\mid X) \rho^\top(A_1 B_1 - \hat A_1 \hat B_1)X] }{} 
\end{align*} 

By the Cauchy-Schwarz inequality, we relate these terms above to the prediction error bounds in \cref{lemma:rrr-bound}, where

\begin{align*}
    \abs{\E\left[ \pi(t \mid X) \rho^\top(A_t B_t - \hat A_t \hat B_t) X\right]}^2 &\leq \underbrace{\E[\pi(t \mid X)^2]}_{<1} \underbrace{\E[((A_t B_t - \hat A_t \hat B_t) X\rho)^2]}_{\leq\frac 1n \|((A_t B_t - \hat A_t \hat B_t) X \rho)\|^2_2}
\end{align*}

Then by \cref{lemma:rrr-bound}, we can bound this term as 
\[\leq \frac 2n \{C_1 \sum_{j>r}d_j^2(A_t B_t X ) + C_2rd_1^2(PE)\}\]

Combining the two terms, we get the final bound
\[ \abs{\hat V^{RR-CV}_{\hat Y}(\pi) - V_Y(\pi)}{} \leq \sqrt{2 \mathrm{log}\left(\frac{2}{\delta}\right) \cdot \frac{\mathrm{Var}[\phi^Y(D^*; \pi)]}{n}} + \frac{2L}{3n}\mathrm{log}\left(\frac{2}{\delta}\right) + 
\sqrt{\frac 2n} 
\{C_1 \sum_{j>r}d_j^2(A_t B_t X ) + C_2rd_1^2(PE)\}^{\frac 12}
\]
 \end{proof}

\begin{proof}[Proof of \Cref{theorem:statanalysis}]
We can define $\hat \pi_n = \arg\mathrm{min}_{\pi \in \Pi} \hat V_{CV}(\pi;\hat Y)$. Applying \Cref{corollary:devboundpopt} twice, once with $\hat \pi_n$ and once with $\pi^*$, and using a uniform upper bound on the variance for all policies in $\Pi$, we obtain  
\[ V_Y(\hat \pi_n) \geq V_Y(\pi^*) - \sqrt{2 \mathrm{log}\left(\frac{2N}{\delta}\right) \cdot \frac{\sup_{\pi \in \Pi}\{ \mathrm{Var}[\phi^Y(D^*; \pi)] \}}{n}} + \frac{2L}{3n}\mathrm{log}\left(\frac{2N}{\delta}\right) + \sqrt{\frac2n} 
\{C_1 \sum_{j>r}d_j^2(A_t B_t X ) + C_2rd_1^2(PE)\}^{\frac 12}
\]
\end{proof}
\section{Additional Experiments and Results}
\subsection{Synthetic Data}

\begin{figure}[ht!]
\centering
\includegraphics[width=0.48\columnwidth]{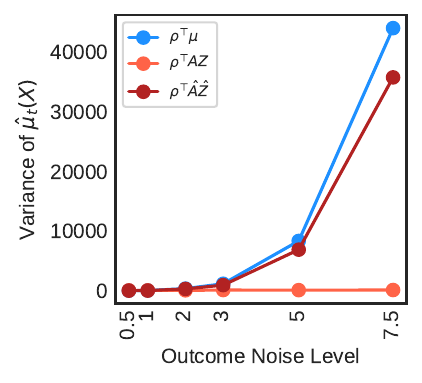}
\includegraphics[width=0.47\columnwidth]{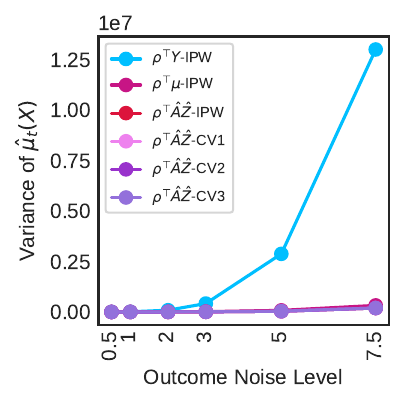}
\caption{\textbf{Left:} Comparison of variances averaged over 100 datasets as the noise in $Y$ increases using the direct estimators for $\rho Y$. \textbf{Right:} Comparison of variances as the noise level in $Y$ increases for the IPW and control variate estimators. \textit{Lower is better.}}
\label{fig:varying_noise}
\end{figure}

\begin{figure*}[ht!]
\centering
\includegraphics[width=0.49\columnwidth]{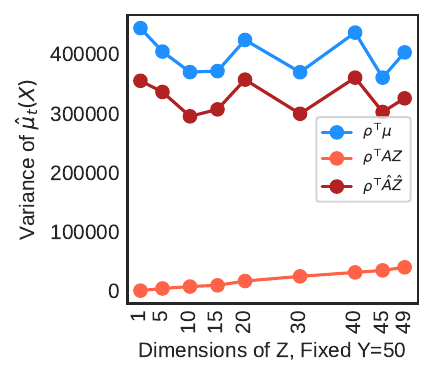}
\includegraphics[width=0.46\columnwidth]{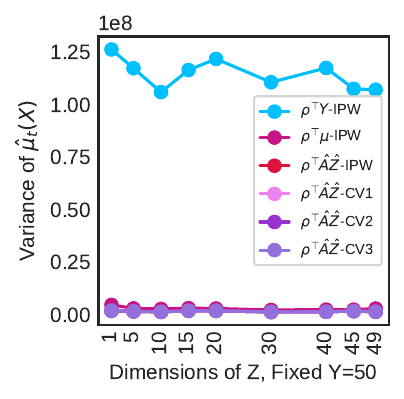}
\includegraphics[width=0.47\columnwidth]{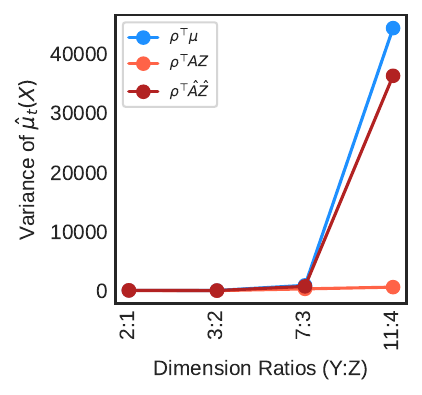}
\includegraphics[width=0.45\columnwidth]{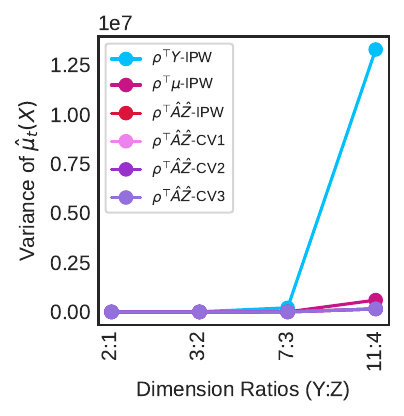}

\caption{\textbf{First Left:} Comparison of variances averaged over 100 datasets as the dimensions of $Y$ and $Z$ increase using the direct estimators for $\rho Y$. The ratio of dimensions is $d_Y/d_Z$ where the dimensions of $Y$ are always greater than $Z$. \textbf{Second Left:} Comparison of variances as the ratio of dimensions of $Y$ to $Z$ is increasing for the IPW and control variate estimators. \textbf{First Right:} Comparison of variances averaged over 100 datasets as the dimensions of $Z$ increase and the dimension of $Y$ remains fixed at $k=50$ for $\rho Y$. \textbf{Second Right:} Comparison of variances as the dimensions of $Z$ increase and the dimensions of $Y$ remain fixed for the IPW and control variate estimators. \textit{Lower is better.}}
\label{fig:vary_dimensions}
\end{figure*}

\Cref{fig:varying_noise} compares the variances as the noise level of the observed outcomes $Y$ increases. \Cref{fig:vary_dimensions} compares the variances as the dimensions of $Y$ and $Z$, where the dimensions of $Y$ are always greater than $Z$. In \Cref{fig:vary_dimensions}, we explore how the variance improvements change in response to varying dimensions of underlying $Z$ and observed $Y$.

In \Cref{fig:po_cvmethod}, we focus on comparing MSE for out-of-sample policy value of policies that are optimal using a scalarized standard doubly-robust method, our control variate with the denoised estimator, and the naive Direct Method (without reduced rank regression). Our method improves upon these other doubly-robust or ablated control variate baselines, especially at small sample sizes. 

Finally, we include a few details about other possible vectors of control variates. (We didn't find any significant differences among these). Instead of choosing $h_t(X) = \hat B_t X$, we could also choose the following: 
\begin{enumerate}
    \item $h_t(X) = \E\left[Y\right]$, no regression control variate
    \item $h_t(X) = \E\left[Y \mid X\right],$ marginalizing over $T$ in the observational data, with regression control variates 
     \item $h_t(X) = \rho^\top\E\left[\hat Z \mid X\right],$ marginalizing over $T$ in the observational data, with regression control variates 
    \item $h_t(X) = \E\left[Y\mid T,X\right]$, with regression control variates
\end{enumerate}
\subsection{Sahel Adaptive Social Protection Program Multi-Country RCT Dataset}

The study is described in \citet{bossuroy2022tackling}\footnote{Data available for download at \url{ https://microdata.worldbank.org/index.php/catalog/4294/get-microdata}}.
We provide additional exploratory analysis for finding heterogeneous treatment effects in the dataset. We complete this analysis using the EconML package hosted by Microsoft. 

\subsubsection{Data cleaning and preprocessing}

\paragraph{Features}
The dataset includes different views on both baseline outcomes and background demographic data, which was collected for a subset of households. We merged the two data sources and kept the households with full information across both baseline and background features. Additionally, after some exploratory analysis, we find that the background demographic data is more predictive, so we narrow down the feature space even further based on availability of both sets of features.

\paragraph{Treatments} Originally, this study had four treatment arms to address the financial and psycho-social dimensions of poverty. The study previously identified 6 productive interventions that would be administered through a countries national cash transfer program. They are (1) needs-based individualized coaching and group-based facilitation (2) community sensitization on aspirations and social norms (3) facilitation of community savings groups (4) micro-entrepreneurship training (5) behavioral skills training (6) one-time cash grant. These interventions were groups into the four treatments arms for impact evaluation. All arms received the regular cash transfer program. All treatment arms ($T_c, T_s, T_f$), not including the control arm, received the core package (savings groups, business training and coaching) which corresponds to interventions (1), (3), and (4). The next treatment arm receives all of the above and the social package $T_s$ which includes life skills training and community sensitization or interventions (2) and (5) respectively. Treatment arm $T_c$ receives all of the above, but instead of the social package, they receive the capital package which includes micro-entrepreneurship training and a cash grant or interventions (4) and (6) respectively. The final treatment arm $T_f$ receives all of the interventions, which includes the core package, the social package and the capital package. For our experiments, we focus on the binary treatment between the control group and those that received the full treatment arm $T_f$. 
\subsubsection{Finding Heterogeneous Treatment Effects} 

\begin{figure*}[ht!]
\centering
\includegraphics[width=0.22\columnwidth]{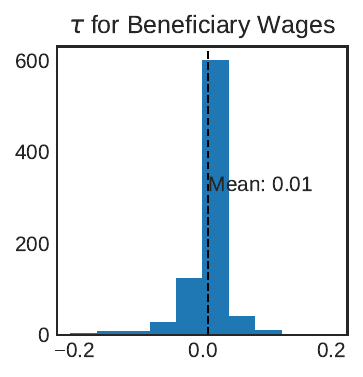}
\includegraphics[width=0.22\columnwidth]{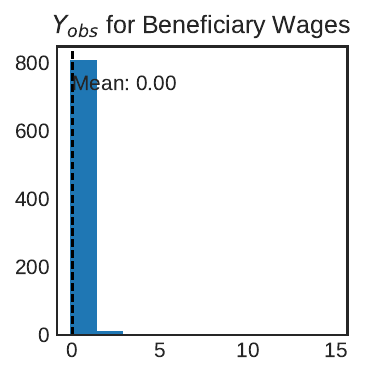}%
\includegraphics[width=0.45\columnwidth]{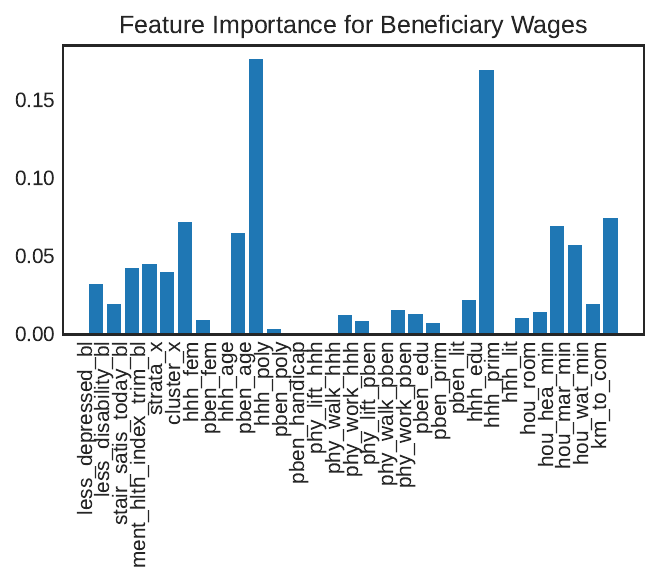}%
\caption{Distribution of treatment effect for beneficiary wages, distribution of observed outcome data in the test set, plot of feature importance from causal forest on training set}
\label{fig:hte_wages}
\end{figure*}

\begin{figure*}[ht!]
\centering
\includegraphics[width=0.22\columnwidth]{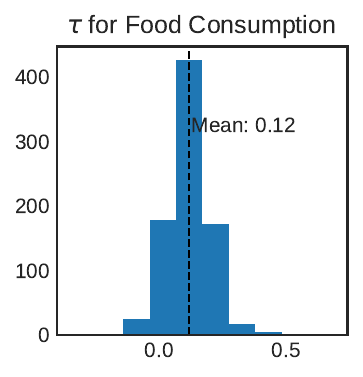}
\includegraphics[width=0.22\columnwidth]{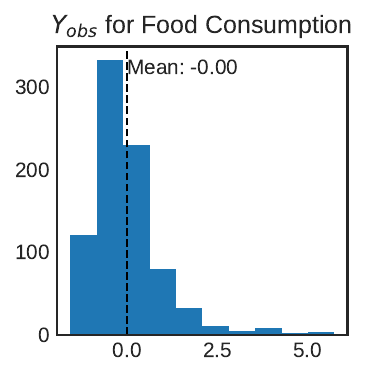}
\includegraphics[width=0.45\columnwidth]{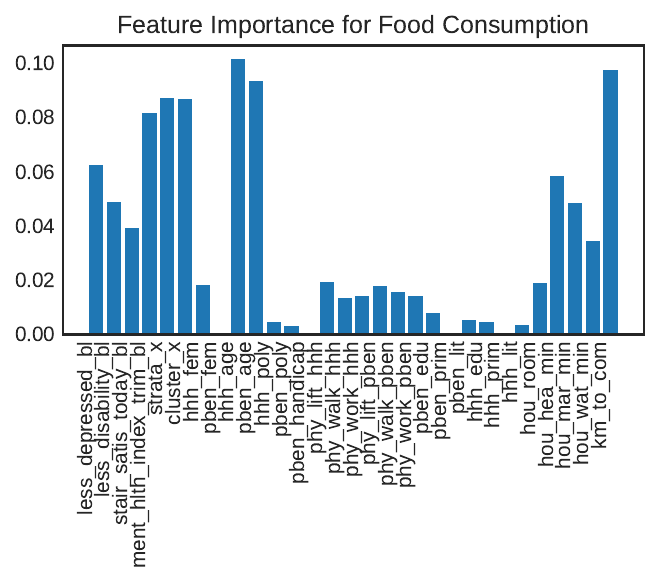}
\caption{Distribution of treatment effect for food consumption, distribution of observed outcome data in the test set, plot of feature importance from causal forest on training set}
\label{fig:hte_consump}
\end{figure*}

\begin{figure*}[ht!]
\centering
\includegraphics[width=0.22\columnwidth]{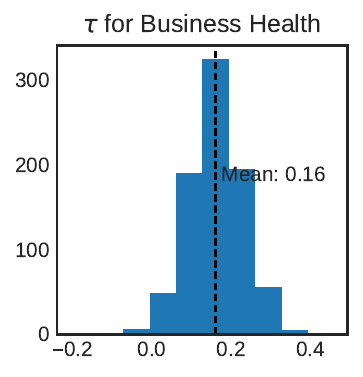}
\includegraphics[width=0.22\columnwidth]{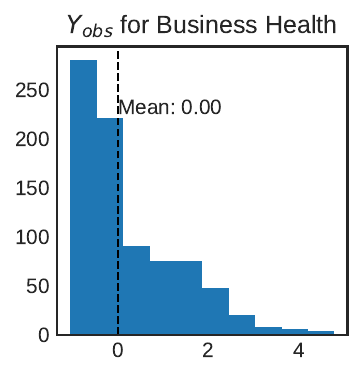}%
\includegraphics[width=0.45\columnwidth]{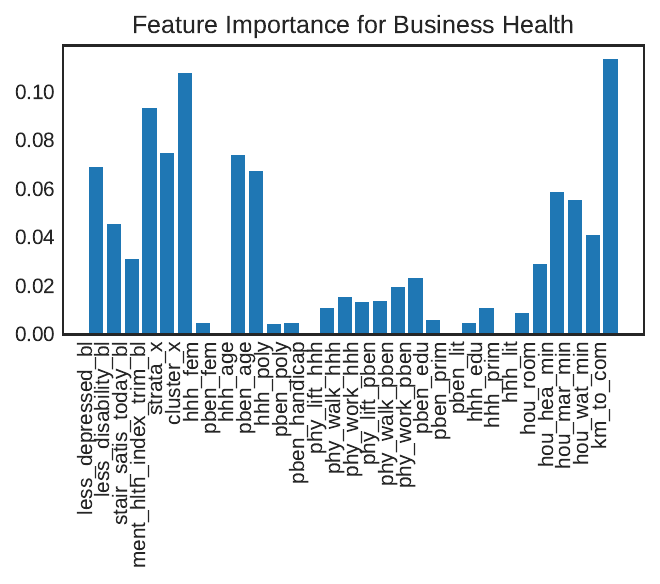}
\caption{Distribution of treatment effect for business health, distribution of observed outcome data in the test set, plot of feature importance from causal forest on training set}
\label{fig:hte_bushealth}
\end{figure*}

\begin{figure*}[ht!]
\centering
\includegraphics[width=0.22\columnwidth]{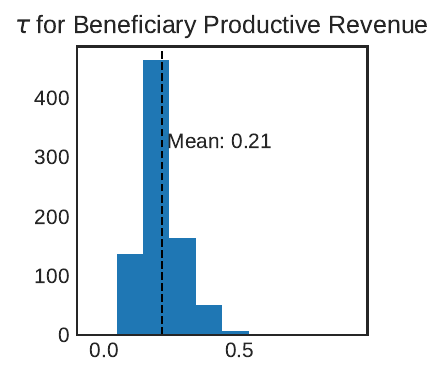}
\includegraphics[width=0.22\columnwidth]{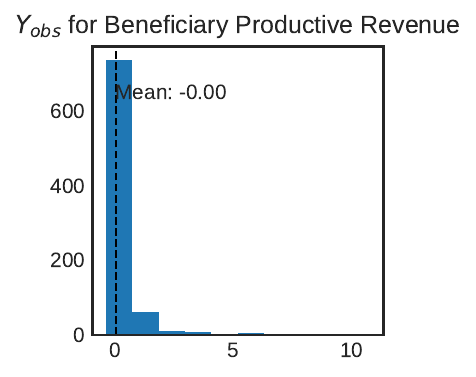}%
\includegraphics[width=0.45\columnwidth]{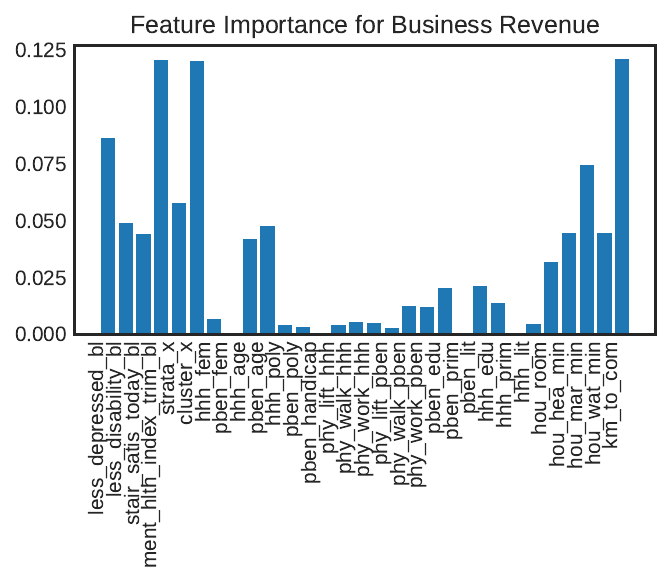}%
\caption{Distribution of treatment effect for business revenue, distribution of observed outcome data in the test set, plot of feature importance from causal forest on training set}
\label{fig:hte_revenue}
\end{figure*}

\begin{figure*}[ht!]
\centering
\includegraphics[width=0.22\columnwidth]{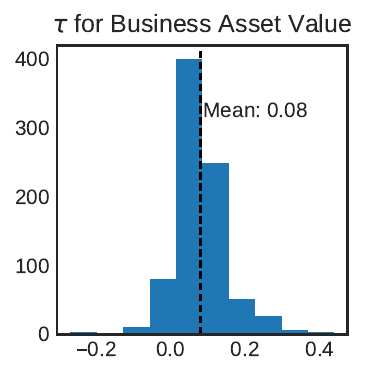}
\includegraphics[width=0.22\columnwidth]{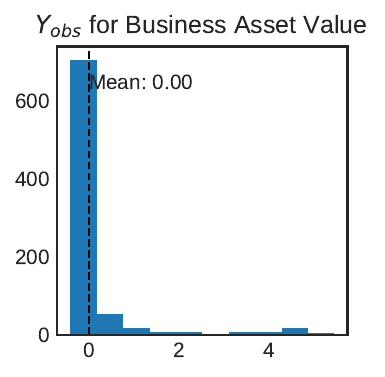}
\includegraphics[width=0.45\columnwidth]{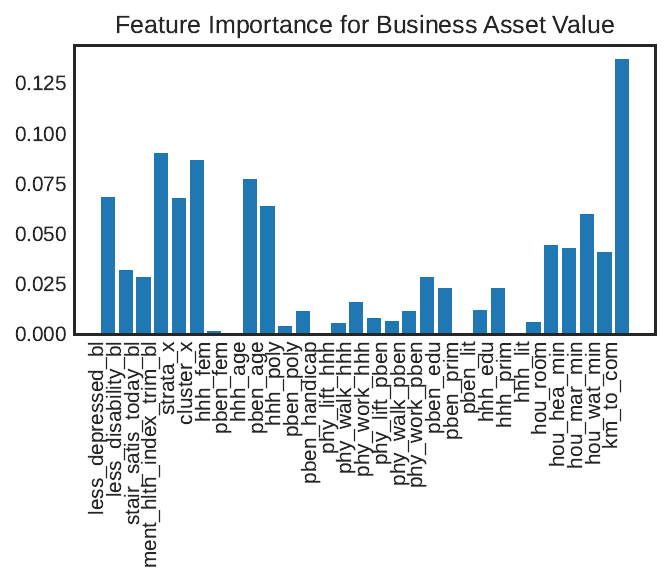}
\caption{Distribution of treatment effect for business asset value, distribution of observed outcome data in the test set, plot of feature importance from causal forest on training set}
\label{fig:hte_assetvalue}
\end{figure*}

\begin{figure*}[ht!]
\centering
\includegraphics[width=0.22\columnwidth]{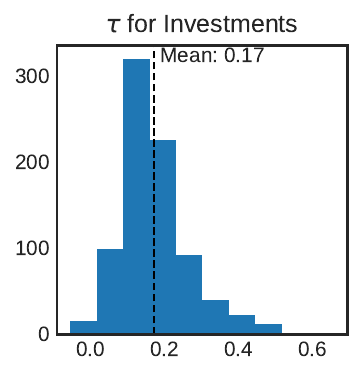}
\includegraphics[width=0.22\columnwidth]{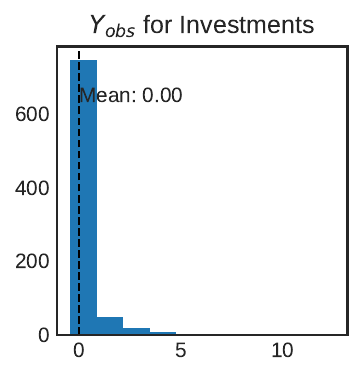}
\includegraphics[width=0.45\columnwidth]{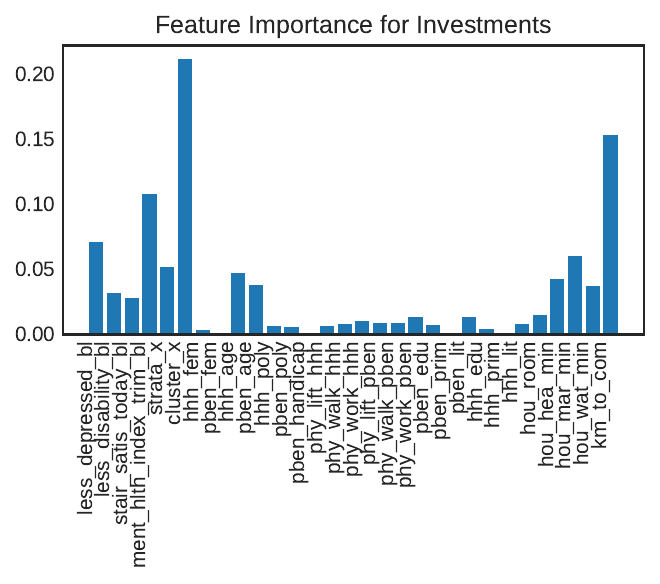}%
\caption{Distribution of treatment effect for investments, distribution of observed outcome data in the test set, plot of feature importance from causal forest on training set}
\label{fig:hte_investments}
\end{figure*}

\begin{figure*}[ht!]
\centering
\includegraphics[width=0.22\columnwidth]{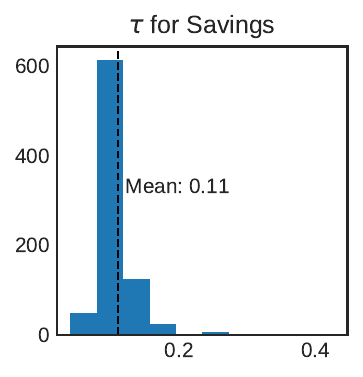}
\includegraphics[width=0.22\columnwidth]{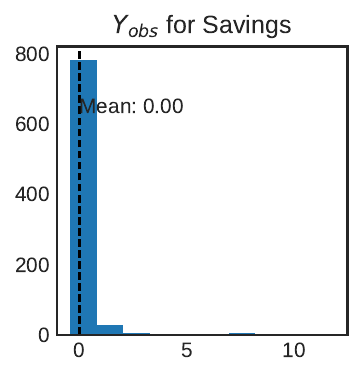}
\includegraphics[width=0.45\columnwidth]{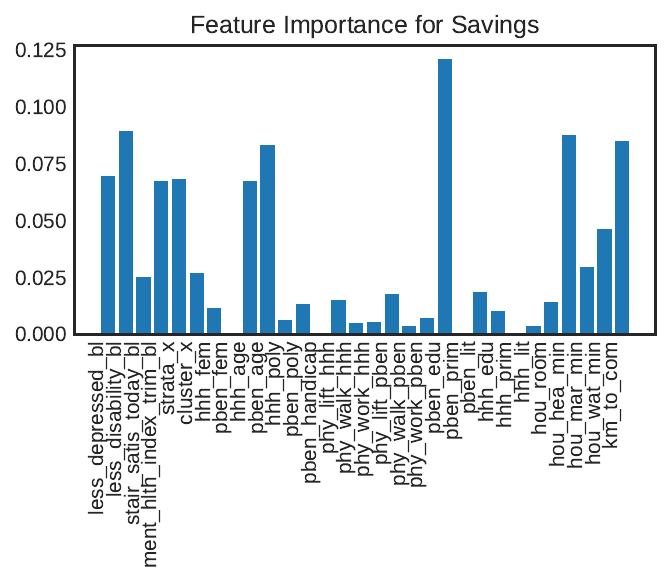}
\caption{Distribution of treatment effect for savings, distribution of observed outcome data in the test set, plot of feature importance from causal forest on training set}
\label{fig:hte_savings}
\end{figure*}

We conduct an exploratory analysis on the seven outcomes of interest to confirm the existence of heterogeneous treatment effects in the Sahel dataset. We provide descriptive information about the estimated causal effect using causal random forests \citep{athey2020pl} and OLS regression with interaction terms between the treatment and covariates. The regression results are summarized in \Cref{hte-regression-table} where we find a few significant interaction terms between the treatment and the features. This is an indication of heterogeneity across the seven outcomes. In \Cref{fig:hte_wages} - \Cref{fig:hte_savings}, the first plot shows the estimates of the causal effect and the second plot shows the distribution of the outcome in the test set.

\begin{table}[t]
\caption{Sahel dataset feature list and feature descriptions.}
\label{feature-list-table}
\vskip 0.1in
\begin{center}
\begin{small}
\resizebox{\columnwidth}{!}{%
\begin{tabular}{lcc}
\toprule
Feature Name & Description  \\
\midrule
less\_depressed\_bl & Depression Baseline: 4 questions from (CESD-R-10), (0-7, recode to 1-4);sum, reversed \\ 
less\_disability\_bl & Disability Baseline: 4 questions from the SRQ-20 (Self-report questionnaire), (recoded from [1-4] to 0/1), (neurotic, stress-related disorders), reversed\\
stair\_satis\_today\_bl & Life Satisfaction: Cantril's latter of life satisfaction (1-10)\\
ment\_hlth\_index\_trim\_bl & One item mental health assessment: Productive beneficiary mental health self-assessment (1-5), standardized\\
hhh\_fem & Household head is female \\
pben\_fem & Productive beneficiary is female \\
hhh\_age & Household head age\\ 
pben\_age & Productive beneficiary age \\
hhh\_poly & Household head is in polygamous relationship (male or female)\\ 
pben\_poly & Productive beneficiary is in polygamous relationship (male or female)\\ 
pben\_handicap & Beneficiary has handicap \\
phy\_lift\_hhh & How difficult is it for the household head to lift a 10 kg bag? \\
phy\_walk\_hhh & How difficult is it for the household head to walk 4 hours without resting? \\
phy\_work\_hhh & How difficult is it for the household head to work in the fields all day? \\
phy\_lift\_pben & How difficult is it for the beneficiary to lift a 10 kg bag? \\
phy\_walk\_pben & How difficult is it for the beneficiary to walk 4 hours without resting? \\ 
phy\_work\_pben & How difficult is it for the beneficiary to work in the fields all day? \\ 
pben\_edu & Productive beneficiary years of education \\
pben\_prim & Productive beneficiary completed primary school \\ 
pben\_lit & Productive beneficiary is literate \\ 
hhh\_edu & Household head years of education \\
hhh\_prim & Household head completed primary school \\
hhh\_lit & Household head is literate \\
hou\_room & Number of rooms in house \\
hou\_hea\_min & Minutes needed to get to nearest health center (usual mode of transport)? \\
hou\_mar\_min & Minutes needed to get to nearest market (usual mode of transport)? \\
hou\_wat\_min & Minutes needed to get to main source of drinking water (usual mode of transport) \\
km\_to\_com & Distance to capital of commune (km) \\
\bottomrule
\end{tabular}}
\end{small}
\end{center}
\vskip -0.1in
\end{table}

\begin{table}[ht!]
\caption{OLS regression table output for each scaled outcome, treatment, and interaction terms with a subset of relevant covariates. The standard errors in parentheses. * $p<.1$, ** $p<.05$, ***$p<.01$}
\label{hte-regression-table}
\begin{center}
\resizebox{\columnwidth}{!}{%
\begin{tabular}{llllllll}
\hline
                        & Beneficiary Wages & Food Consumption & Business Health & Beneficiary Productive Revenue & Business Asset Value & Investments & Savings  \\
\hline
Intercept               & 0.3624*            & 0.1321       & -0.4559**        & 0.3673*            & -0.0356              & 0.1010                & -0.3024         \\
                        & (0.2131)           & (0.2120)     & (0.2148)         & (0.2064)           & (0.2118)             & (0.2071)              & (0.2103)        \\
treatment               & 0.1178             & -0.1631      & 0.1452           & 0.2315             & 0.1015               & -0.2885               & 0.2275          \\
                        & (0.3163)           & (0.3121)     & (0.3174)         & (0.3063)           & (0.3146)             & (0.3075)              & (0.3122)        \\
hhh\_age                & -0.0017            & -0.0003      & 0.0031           & 0.0001             & 0.0037**             & 0.0016                & 0.0032*         \\
                        & (0.0019)           & (0.0018)     & (0.0019)         & (0.0018)           & (0.0018)             & (0.0018)              & (0.0018)        \\
treatment:hhh\_age      & 0.0038             & 0.0019       & -0.0004          & 0.0007             & -0.0025              & -0.0023               & -0.0036         \\
                        & (0.0026)           & (0.0026)     & (0.0027)         & (0.0025)           & (0.0026)             & (0.0026)              & (0.0026)        \\
pben\_age               & -0.0012            & 0.0018       & 0.0048**         & 0.0027             & -0.0040**            & 0.0006                & -0.0006         \\
                        & (0.0019)           & (0.0019)     & (0.0022)         & (0.0019)           & (0.0019)             & (0.0019)              & (0.0019)        \\
treatment:pben\_age     & -0.0073***         & -0.0047*     & -0.0059**        & -0.0009            & 0.0041               & 0.0019                & 0.0015          \\
                        & (0.0027)           & (0.0027)     & (0.0029)         & (0.0027)           & (0.0027)             & (0.0027)              & (0.0027)        \\
hhh\_fem                & 0.0137             & 0.2092***    & 0.0524           & 0.0568             & -0.0794              & 0.1070                & 0.0439          \\
                        & (0.0725)           & (0.0711)     & (0.0746)         & (0.0702)           & (0.0720)             & (0.0705)              & (0.0715)        \\
treatment:hhh\_fem      & 0.2028**           & -0.1137      & -0.0660          & -0.0578            & -0.1581              & -0.2036**             & -0.0179         \\
                        & (0.1016)           & (0.0994)     & (0.1027)         & (0.0984)           & (0.1006)             & (0.0988)              & (0.0999)        \\
pben\_fem               & -0.2608            & -0.2878      & -0.0207          & -0.6038***         & -0.0318              & -0.3847**             & 0.0823          \\
                        & (0.1926)           & (0.1921)     & (0.1927)         & (0.1865)           & (0.1917)             & (0.1872)              & (0.1903)        \\
treatment:pben\_fem     & 0.0380             & 0.5083*      & 0.4399           & 0.2023             & 0.0754               & 0.7009**              & 0.2081          \\
                        & (0.2927)           & (0.2890)     & (0.2935)         & (0.2835)           & (0.2914)             & (0.2846)              & (0.2892)        \\
hou\_hea\_min           & -0.0004            & 0.0004       & 0.0008           & 0.0002             & -0.0011*             & 0.0000                & 0.0002          \\
                        & (0.0006)           & (0.0006)     & (0.0006)         & (0.0006)           & (0.0006)             & (0.0006)              & (0.0006)        \\
treatment:hou\_hea\_min & -0.0006            & -0.0009      & -0.0014*         & -0.0023***         & 0.0006               & -0.0006               & -0.0009         \\
                        & (0.0008)           & (0.0008)     & (0.0008)         & (0.0008)           & (0.0008)             & (0.0008)              & (0.0008)        \\
hou\_mar\_min           & 0.0002             & -0.0007*     & -0.0005          & -0.0005            & 0.0003               & -0.0004               & -0.0001         \\
                        & (0.0004)           & (0.0004)     & (0.0004)         & (0.0004)           & (0.0004)             & (0.0004)              & (0.0004)        \\
treatment:hou\_mar\_min & 0.0003             & -0.0002      & 0.0007           & 0.0003             & -0.0003              & 0.0008                & -0.0011*        \\
                        & (0.0006)           & (0.0006)     & (0.0006)         & (0.0006)           & (0.0006)             & (0.0006)              & (0.0006)        \\
hou\_wat\_min           & -0.0010            & 0.0000       & -0.0011          & -0.0011            & 0.0017               & 0.0023**              & -0.0010         \\
                        & (0.0011)           & (0.0011)     & (0.0011)         & (0.0011)           & (0.0011)             & (0.0011)              & (0.0011)        \\
treatment:hou\_wat\_min & -0.0002            & 0.0016       & 0.0009           & -0.0017            & -0.0052***           & -0.0039**             & 0.0000          \\
                        & (0.0017)           & (0.0017)     & (0.0017)         & (0.0016)           & (0.0017)             & (0.0017)              & (0.0017)        \\
R-squared               & 0.0099             & 0.0174       & 0.0340           & 0.0366             & 0.0120               & 0.0324                & 0.0150          \\
R-squared Adj.          & 0.0065             & 0.0140       & 0.0305           & 0.0333             & 0.0086               & 0.0291                & 0.0117          \\
\hline
\end{tabular}}
\end{center}
\end{table}

\paragraph{Data transformations for reduced rank regression}
We log-transform the outcomes: beneficiary wages, beneficiary productive revenue, business asset value, investments, and savings, as is common in econometrics:  histograms in the appendix illustrate the right-tailed distribution. Lastly, we tune the number of factors that we estimate for $\hat Y$, though this ends up quite similar to our results in the main paper with two factors. 

\begin{table*}[t!]
\caption{Out-of-sample evaluation of optimized policy value (\textbf{higher} is better). Rows indicate policies optimizing different estimates on the training set. Columns indicate different evaluation methods on the test set. Estimated factors $\hat Z$ = 3. We report the mean $\pm$ standard deviation. Bold is best (within-column), italic second-best.}
\label{model-robust-table1}
\vskip 0.1in
\begin{center}
\begin{small}
\begin{tabular}{lcccc}
\toprule
$\pi^*$& Self  & $\rho^\top \mu^{RR-DM}$   & $\rho^\top Y^{IPW}$   & $\rho^\top Y^{DR}$  
\\
\midrule
 $\rho^\top \mu^{DM}$ & 0.101 $\pm$ 1.836   & 0.083 $\pm$ 1.548          & 0.180 $\pm$ 11.959  & 0.149 $\pm$ 11.851 
    
 \\
 $\rho^\top \mu^{RR-DM}$  & 0.168 $\pm$ 1.609   & \textbf{0.168} $\pm$ 1.609          & \textbf{0.363} $\pm$ 12.557  & \textbf{0.315} $\pm$ 12.437  
   
 \\
 $\rho^\top Y^{IPW}$ & -0.224 $\pm$ 10.120 & -0.255 $\pm$ 1.672         & -0.224 $\pm$ 10.120 & -0.199 $\pm$ 10.098 
  
 \\
 $\rho^\top \mu^{RR-IPW}$ & 0.195 $\pm$ 2.376   & \textit{0.151} $\pm$ 1.639          & \textit{0.321} $\pm$ 12.348  & \textit{0.278} $\pm$ 12.240  
   
 \\
 $\rho^\top Y^{DR}$ & 0.008 $\pm$ 10.275  & -0.208 $\pm$ 1.682         & -0.025 $\pm$ 10.355 & 0.008 $\pm$ 10.275  

 \\
 $\rho^\top \mu^{RR-CV}$  & 0.194 $\pm$ 2.381   & \textit{0.151} $\pm$ 1.639          & \textit{0.321} $\pm$ 12.349  & \textit{0.278} $\pm$ 12.241 
   
 \\
\bottomrule
\end{tabular}
\end{small}
\end{center}
\vskip -0.1in
\end{table*}

\begin{table*}[t!]
\caption{Out-of-sample evaluation of optimized policy value (\textbf{higher} is better). Rows indicate policies optimizing different estimates on the training set. Columns indicate different evaluation methods on the test set. Estimated factors $\hat Z$ = 4. We report the mean $\pm$ standard deviation. Bold is best (within-column), italic second-best.}
\label{model-robust-table2}
\vskip 0.1in
\begin{center}
\begin{small}
\begin{tabular}{lcccc}
\toprule
$\pi^*$& Self  & $\rho^\top \mu^{RR-DM}$   & $\rho^\top Y^{IPW}$   & $\rho^\top Y^{DR}$  
\\
\midrule
 $\rho^\top \mu^{DM}$ & 0.101 $\pm$ 1.836   & 0.085 $\pm$ 1.582          & \textit{0.180} $\pm$ 11.959  & \textit{0.148} $\pm$ 11.835 
   
 \\
 $\rho^\top \mu^{RR-DM}$  & 0.108 $\pm$ 1.618   & \textbf{0.108} $\pm$ 1.618          & \textit{0.180} $\pm$ 12.166  & 0.145 $\pm$ 11.994  
   
 \\
 $\rho^\top Y^{IPW}$ & -0.224 $\pm$ 10.120 & -0.304 $\pm$ 1.691         & -0.224 $\pm$ 10.120 & -0.216 $\pm$ 10.100 
   
 \\
 $\rho^\top \mu^{RR-IPW}$ & 0.124 $\pm$ 2.461   & \textit{0.103} $\pm$ 1.628          & \textbf{0.211} $\pm$ 12.313  & \textbf{0.190} $\pm$ 12.143  
   
 \\
 $\rho^\top Y^{DR}$ & -0.056 $\pm$ 10.149 & -0.265 $\pm$ 1.710         & -0.077 $\pm$ 10.228 & -0.056 $\pm$ 10.149 
 \\
 $\rho^\top \mu^{RR-CV}$  & 0.124 $\pm$ 2.468   & \textit{0.103} $\pm$ 1.628          & \textbf{0.211} $\pm$ 12.313  & \textbf{0.190} $\pm$ 12.144  
 \\
\bottomrule
\end{tabular}
\end{small}
\end{center}
\vskip -0.1in
\end{table*}

\begin{table*}[t!]
\caption{Out-of-sample evaluation of optimized policy value (\textbf{higher} is better). Rows indicate policies optimizing different estimates on the training set. Columns indicate different evaluation methods on the test set. Estimated factors $\hat Z$ = 5. We report the mean $\pm$ standard deviation. Bold is best (within-column), italic second-best.}
\label{model-robust-table3}
\vskip 0.1in
\begin{center}
\begin{small}
\begin{tabular}{lcccc}
\toprule
$\pi^*$& Self  & $\rho^\top \mu^{RR-DM}$   & $\rho^\top Y^{IPW}$   & $\rho^\top Y^{DR}$  
\\
\midrule
 $\rho^\top \mu^{DM}$ & 0.101 $\pm$ 1.836   & \textbf{0.128} $\pm$ 1.723          & \textbf{0.180} $\pm$ 11.959  & \textbf{0.160} $\pm$ 11.772  
   
 \\
 $\rho^\top \mu^{RR-DM}$  & 0.117 $\pm$ 1.765   & \textit{0.117} $\pm$ 1.765          & \textit{0.149} $\pm$ 11.928  & 0.125 $\pm$ 11.719  
   
 \\
 $\rho^\top Y^{IPW}$ & -0.224 $\pm$ 10.120 & -0.326 $\pm$ 1.822         & -0.224 $\pm$ 10.120 & -0.217 $\pm$ 10.103 
   
 \\
 $\rho^\top \mu^{RR-IPW}$ & 0.081 $\pm$ 2.498   & 0.079 $\pm$ 1.733          & 0.129 $\pm$ 11.632  & \textit{0.127} $\pm$ 11.458  
   
 \\
 $\rho^\top Y^{DR}$ & 0.050 $\pm$ 10.445  & -0.278 $\pm$ 1.814         & 0.028 $\pm$ 10.546  & 0.050 $\pm$ 10.445  
 \\
 $\rho^\top \mu^{RR-CV}$  & 0.082 $\pm$ 2.505   & 0.079 $\pm$ 1.733          & 0.130 $\pm$ 11.632  & \textit{0.127} $\pm$ 11.459  
 \\
\bottomrule
\end{tabular}
\end{small}
\end{center}
\vskip -0.1in
\end{table*}

\begin{table*}[t!]
\caption{Out-of-sample evaluation of optimized policy value (\textbf{higher} is better). Rows indicate policies optimizing different estimates on the training set. Columns indicate different evaluation methods on the test set. Estimated factors $\hat Z$ = 6. We report the mean $\pm$ standard deviation. Bold is best (within-column), italic second-best.}
\label{model-robust-table4}
\vskip 0.1in
\begin{center}
\begin{small}
\begin{tabular}{lcccc}
\toprule
$\pi^*$& Self  & $\rho^\top \mu^{RR-DM}$   & $\rho^\top Y^{IPW}$   & $\rho^\top Y^{DR}$  
\\
\midrule
 $\rho^\top \mu^{DM}$ & 0.101 $\pm$ 1.836   & \textbf{0.090} $\pm$ 1.774          & \textbf{0.180} $\pm$ 11.959  & \textbf{0.151} $\pm$ 11.767 
   
 \\
 $\rho^\top \mu^{RR-DM}$  & 0.081 $\pm$ 1.777   & \textit{0.081} $\pm$ 1.777          & \textit{0.168} $\pm$ 11.919  & \textit{0.137} $\pm$ 11.733  
   
 \\
 $\rho^\top Y^{IPW}$  & -0.224 $\pm$ 10.120 & -0.285 $\pm$ 1.835         & -0.224 $\pm$ 10.120 & -0.212 $\pm$ 10.119 
   
 \\
 $\rho^\top \mu^{RR-IPW}$ & 0.068 $\pm$ 2.432   & 0.047 $\pm$ 1.750          & 0.124 $\pm$ 11.859  & 0.103 $\pm$ 11.723  
   
 \\
 $\rho^\top Y^{DR}$ & -0.038 $\pm$ 10.165 & -0.239 $\pm$ 1.851         & -0.069 $\pm$ 10.240 & -0.038 $\pm$ 10.165 
 \\
 $\rho^\top \mu^{RR-CV}$  & 0.067 $\pm$ 2.442   & 0.047 $\pm$ 1.750          & 0.124 $\pm$ 11.860  & 0.103 $\pm$ 11.723  
 \\
\bottomrule
\end{tabular}
\end{small}
\end{center}
\vskip -0.1in
\end{table*}

\end{document}